\documentclass{article}
\usepackage[preprint]{neurips_2026}

\usepackage[colorlinks=true, citecolor=hanpurple, urlcolor=hanpurple, linkcolor=fireenginered]{hyperref}  
\usepackage{amsmath,amssymb,amsfonts,amsthm}
\usepackage{graphicx}
\usepackage{booktabs}
\usepackage{hyperref}
\usepackage{xcolor}
\usepackage{enumitem}
\usepackage{subcaption}
\usepackage{pifont}
\usepackage{multirow}
\usepackage{nicefrac}
\usepackage{algorithm}
\usepackage{algorithmic}
\usepackage{longtable}
\usepackage{colortbl}
\usepackage{xspace}
\usepackage{wrapfig}
\usepackage{listings}
\usepackage[most]{tcolorbox}
\usepackage{titletoc}
\usepackage{pifont}

\newcommand\DoToC{%
  \startcontents
  \begingroup
  \setcounter{tocdepth}{2}
  \printcontents{}{1}{\textbf{Contents of Appendix}\vskip3pt\hrule\vskip5pt}
  \vskip3pt\hrule\vskip5pt
  \endgroup
}

\definecolor{lightblue}{RGB}{220,235,252}
\definecolor{lightgreen}{RGB}{220,252,225}
\definecolor{lightyellow}{RGB}{255,250,220}
\definecolor{lightgray}{RGB}{245,245,245}
\definecolor{bestcolor}{RGB}{230,245,230}
\definecolor{deeppurple}{RGB}{60,52,137}
\definecolor{hanpurple}{rgb}{0.32, 0.09, 0.98}
\definecolor{divoff}{HTML}{D97706}
\definecolor{divon}{HTML}{4F46E5}
\definecolor{fireenginered}{rgb}{0.81, 0.09, 0.13}

\newcommand{\sys}{\texttt{Harness-1}\xspace}
\newcommand{\wm}{\textsc{WorkingMemory}\xspace}

\newcommand{\verify}{\texttt{verify}\xspace}
\newcommand{\curate}{\texttt{curate}\xspace}

\newcommand{\fourstarwhite}{\ding{71}} 

\newtcolorbox{keybox}[1][]{
  colback=lightblue!60,
  colframe=blue!50!black,
  boxrule=0.5pt,
  arc=2pt,
  left=6pt,right=6pt,top=4pt,bottom=4pt,
  #1
}

\newtcolorbox{promptbox}[2][]{
  enhanced,
  breakable,
  colback=lightgray!35,
  colframe=black!45,
  boxrule=0.4pt,
  arc=2pt,
  left=8pt,right=8pt,top=6pt,bottom=6pt,
  title={#2},
  fonttitle=\sffamily\small\bfseries,
  fontupper=\small\sffamily,
  #1
}

\definecolor{toolFanout}{HTML}{2BA6A6}
\definecolor{toolSearch}{HTML}{4C78A8}
\definecolor{toolGrep}{HTML}{7B61A8}
\definecolor{toolRead}{HTML}{F28E2B}
\definecolor{toolReview}{HTML}{B07AA1}
\definecolor{toolCurate}{HTML}{59A14F}
\definecolor{toolVerify}{HTML}{E15759}
\definecolor{toolStop}{HTML}{8D8D8D}
\definecolor{toolNeutral}{HTML}{5B6770}
\definecolor{docred}{RGB}{180,55,55}
\definecolor{docblue}{RGB}{45,95,170}
\newcommand{\CaseTool}[2]{\begingroup\setlength{\fboxsep}{1.6pt}\colorbox{#1!12}{\textcolor{#1!75!black}{\texttt{#2}}}\endgroup}
\lstdefinestyle{trajstyle}{
  basicstyle=\sffamily\footnotesize,
  breaklines=true,
  breakatwhitespace=false,
  breakautoindent=false,
  breakindent=0pt,
  columns=fullflexible,
  keepspaces=false,
  showstringspaces=false,
  xleftmargin=0.15em,
  xrightmargin=0.15em
}
\newtcblisting{trajectorylisting}[3][]{
  enhanced,
  breakable,
  listing only,
  listing options={style=trajstyle},
  colback=#2!7,
  colframe=#2!80!black,
  colbacktitle=#2!75!black,
  coltitle=white,
  boxrule=0.6pt,
  arc=2pt,
  left=7pt,right=7pt,top=5pt,bottom=5pt,
  title={#3},
  fonttitle=\sffamily\small\bfseries,
  #1
}
\newtcolorbox{casenote}[2][]{
  enhanced,
  breakable,
  colback=#2!8,
  colframe=#2!80!black,
  colbacktitle=#2!75!black,
  coltitle=white,
  boxrule=0.6pt,
  arc=2pt,
  left=8pt,right=8pt,top=6pt,bottom=6pt,
  fonttitle=\sffamily\small\bfseries,
  fontupper=\small\sffamily,
  #1
}

\title{Harness-1: Reinforcement Learning for Search Agents with State-Externalizing Harnesses}
\author{
  Pengcheng Jiang \hspace{0.5em} Zhiyi Shi \hspace{0.3em} Kelly Hong$^\dagger$\textsuperscript{\fourstarwhite} \hspace{0.3em} Xueqiang Xu \hspace{0.3em} Jiashuo Sun \\
  \textbf{Jimeng Sun} \hspace{0.3em} \textbf{Hammad Bashir}\textsuperscript{\fourstarwhite} \hspace{0.3em} \textbf{Jiawei Han}\\
  University of Illinois at Urbana-Champaign \quad $^\dagger$UC Berkeley \quad \textsuperscript{\fourstarwhite}Chroma \\
}

\begin{document}
\maketitle

\begin{abstract}
Search agents are often trained as policies over growing transcripts: the model must decide how to search while also remembering what it has seen, which evidence is useful, which constraints remain open, and which claims have actually been checked. We argue that this formulation puts too much routine state management inside the policy: reinforcement learning is forced to optimize both semantic search decisions and recoverable bookkeeping that the environment can maintain more reliably.
\
We introduce \sys, a 20B search agent (retrieval subagent) trained with reinforcement learning inside a stateful search harness. The harness maintains environment-side working memory, including a candidate pool, an importance-tagged curated set, compact evidence links, verification records, compressed and deduplicated observations, and budget-aware context rendering. The policy retains the semantic decisions: what to search, which documents to keep or discard, what to verify, and when to stop.
\
Across eight retrieval benchmarks spanning web, finance, patents, and multi-hop QA, \sys achieves 0.730 average curated recall, outperforming the next strongest open search subagent by +11.4 points and remaining competitive with much larger frontier-model searchers. Its gains are especially strong on held-out transfer benchmarks, suggesting that reinforcement learning over explicit search state can produce retrieval behaviors that generalize beyond the training domains. Our code is available at \url{https://github.com/pat-jj/harness-1}.
\end{abstract}

\vspace{-1em}
\begin{figure}[h]
\centering
\includegraphics[width=\textwidth]{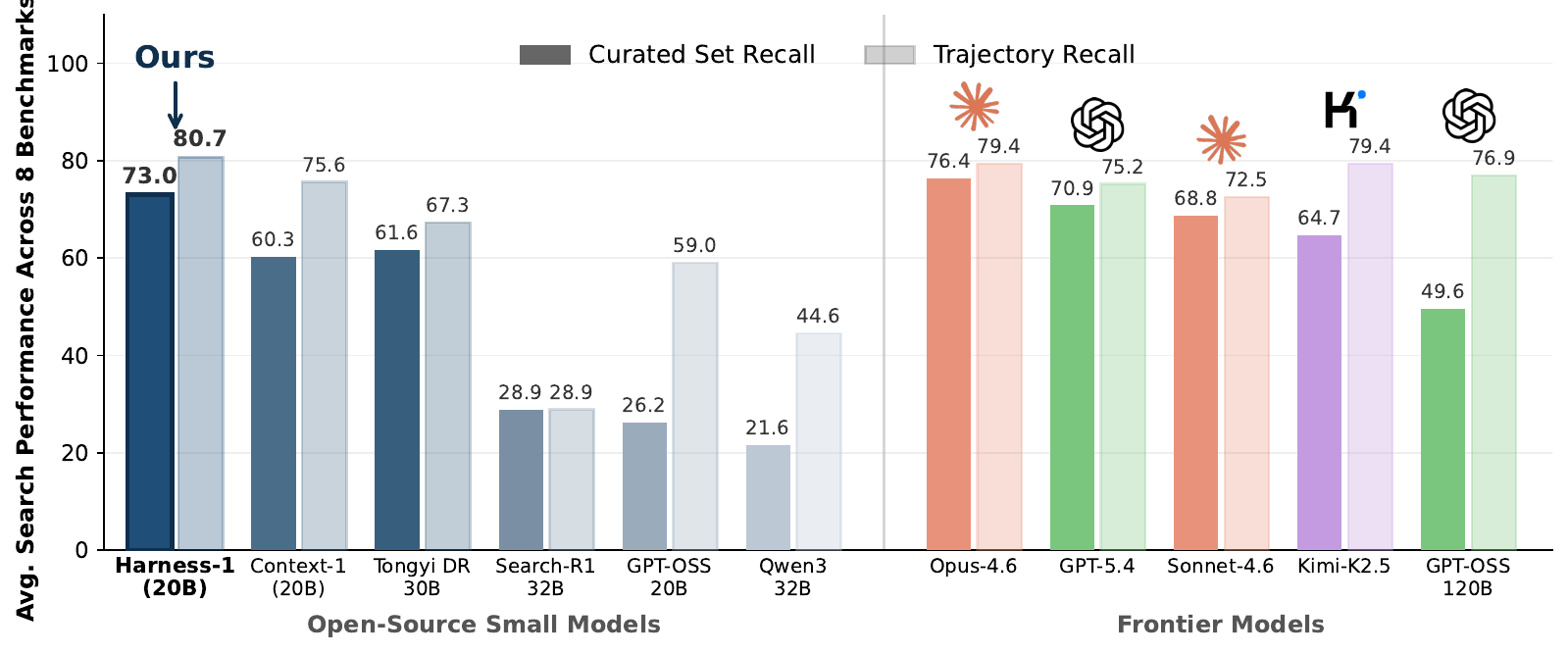}
\caption{\textbf{Averaged performance across eight challenging search benchmarks.}
Each method reports: Curated Set Recall and Trajectory Recall
(documents encountered anywhere in the episode). \sys is our 20B open search
agent trained with a stateful harness; it substantially improves over open
search agents and remains competitive with much larger frontier-model
searchers.}
\label{fig:teaser}
\end{figure}

\section{Introduction}
\label{sec:intro}

Search agents are usually described as language models that call retrieval
tools: given a question, they issue queries, read returned evidence, decide what
is still missing, and return documents to a downstream answerer. This view
follows a broad line of work on reasoning-and-acting agents, iterative
retrieval, active retrieval, and tool-use training
\citep{yao2023react,trivedi2023interleaving,jiang2023active,qin2023toolllm,jiang2026adaptationagenticaisurvey}.
It captures the visible behavior of the agent, but not the search state that
must be built along the way. A successful multi-turn search episode requires
the agent to remember which documents have been seen, which candidates are worth
keeping, which constraints remain uncovered, which entities connect separate
pieces of evidence, and which claims have actually been checked against source
text.

This distinction becomes especially important for reinforcement learning.
Recent work has shown that LLMs can be trained to interact with search engines
and retrieval systems through RL, improving query generation, multi-turn search,
and downstream retrieval utility
\citep{jiang2025deepretrieval,jin2025searchr1,jiang2025s3,wang2025stepsearch}. Yet most
retrieval-agent training still asks the model to learn two different things at
once: semantic search behavior and routine state management. As the transcript
grows, the policy must reconstruct the useful state of the search from raw
observations. This is inefficient, and it can make learning poorly conditioned.
Hard queries may give rollouts nearly identical empty-set rewards; larger tool
vocabularies may collapse to repeated search calls; and cross-document
structure may be present in the transcript but too diffuse to use reliably. When the final curated set is empty or wrong, the reward often does not reveal
whether the failure came from bad search, forgotten evidence, missing
verification, or poor curation.

\begin{keybox}
\textbf{Stateful cognitive offloading.}
A retrieval policy should make semantic decisions over explicit search state:
what to search, which documents to keep, what claim to verify, and when to stop.
The harness should maintain the recoverable state around those decisions:
candidate pools, curated evidence, cross-document links, verification records,
and context-budget summaries. This gives RL a stable interface for improving
search behavior, rather than asking the model to rediscover bookkeeping from an
append-only transcript.
\end{keybox}

This view changes what tool orchestration means. Prior work on agent-computer
interfaces and software-engineering agents has shown that the interface around
a language model can strongly shape agent behavior and performance
\citep{yang2024sweagent,xia2025demystifying}. In retrieval, recent systems have
begun to treat context management as part of the search process, for example
through context pruning or self-editing search \citep{bashir2026context1}. In
\sys, the interface itself is part of the method. The agent acts over a
persistent retrieval state maintained by the environment, and each observation
renders that state back to the policy. The policy still decides what to search,
which documents to promote, what to verify, and when the evidence is sufficient.
The harness carries the state maintenance that would otherwise consume the
policy's context and optimization budget.

We instantiate this principle in \sys, a 20B search agent built on
\texttt{gpt-oss-20b}~\citep{agarwal2025gpt}. At each turn, the model observes recent tool results
together with a rendered \wm containing the candidate pool, curated set, search
history, evidence graph summary, verification state, and context-budget marker.
The model's actions edit this state rather than merely extending a transcript:
\curate adds, removes, and importance-tags documents; \verify checks
policy-written claims against remembered documents; and
\texttt{review\_docs} re-renders previously seen documents without a new corpus
call. Search observations are compressed and deduplicated before reaching the
prompt. The same state renderer is used for teacher data generation, SFT replay,
RL rollout, and evaluation.

\begin{figure}[t]
\centering
\includegraphics[width=\textwidth]{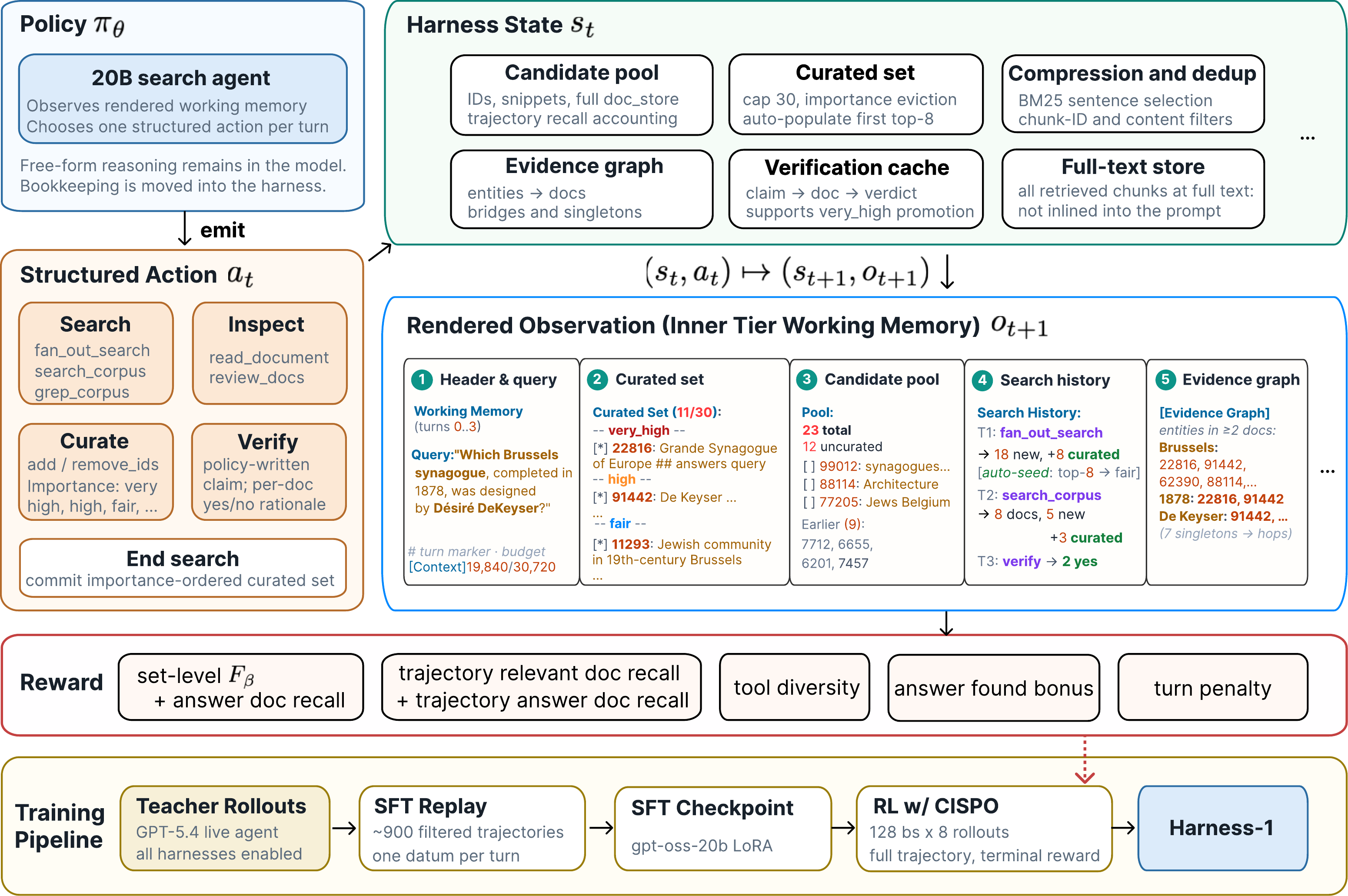}
\caption{\textbf{Overview of \sys.} The policy makes semantic decisions over
search, inspection, curation, verification, and termination. The harness
maintains the state around those decisions: document pool, importance-tagged
curated set, evidence graph, verification records, compression and
deduplication, result summaries, and budget markers. The same state interface is
used for teacher rollouts, SFT replay, CISPO RL, and evaluation.}
\label{fig:harness}
\end{figure}

A richer harness does not automatically yield an environment suitable for policy training. If every rollout starts with an
empty curated set, early rewards can be indistinguishable. If derived state is
too verbose, it competes with evidence for context. If reward only values
discovery, the policy can ignore curation and verification and converge to
repeated search. We therefore identify three requirements for stateful search
harnesses: \textbf{warm-started curation}, implemented by auto-seeding a
tentative curated set from the first successful search; \textbf{compact
derived-state rendering}, implemented through importance tags, evidence-graph
summaries, and verification records; and \textbf{diversity-preserving
incentives}, which encourage a rhythm of searching, curating, reviewing, and
verifying.

We evaluate \sys on eight retrieval benchmarks spanning web, finance, patents,
and long-context multi-hop search. It reaches $0.730$ average curated recall and
improves downstream answer accuracy when its curated set is passed to frozen
frontier generators. The gains are not confined to the training domains: the
same operations over explicit search state improve retrieval across specialized
corpora, open-web tasks, and held-out multi-hop QA benchmarks.

\textbf{Contributions.}
First, we formulate harness design as a central problem in retrieval-agent RL
and introduce stateful cognitive offloading as a principle for assigning search
state between policy and environment. Second, we introduce \sys, a 20B search agent with an importance-tagged curated set, compact evidence graph,
verification records, compressed and deduplicated observations, and
budget-aware context rendering. Third, we identify three trainability
requirements for stateful search harnesses: warm-started curation, compact
derived-state rendering, and diversity-preserving incentives. Fourth, we show
that stateful harnessing improves retrieval quality and modular RAG answer
accuracy across eight benchmarks. We will release the model weights, harness
code, data-generation pipeline, and RL recipe.

\section{\sys}
\label{sec:method}

\sys is a search agent that produces a ranked set of documents for a downstream
answering model. It runs inside a state-machine harness centered on a
per-episode \wm. At each turn, the harness renders compact search state together
with recent actions and observations; the model emits one structured Harmony
action; and the harness executes the action, updates the state, and renders the
next observation.

The harness does not make retrieval decisions for the model. Instead, it
maintains the recoverable state around those decisions. The policy still decides
what to search, which documents to inspect or keep, what claims to verify, and
when to stop. Each action applies a transition
$(s_t,a_t)\mapsto(s_{t+1},o_{t+1})$ over the harness state summarized in
Table~\ref{tab:state-offload}. This is the sense in which the harness is
\emph{stateful}: tool outputs do not merely produce text for the next prompt;
they update persistent search state that future actions can inspect, edit, or
verify.

\begin{table}[t]
\centering
\small
\caption{State notation and offloaded bookkeeping in the \sys harness.
The harness maintains and renders state; the policy retains the semantic
choices over that state.}
\label{tab:state-offload}
\resizebox{\textwidth}{!}{
\begin{tabular}{lll}
\toprule
\textbf{State} & \textbf{Harness-maintained bookkeeping} & \textbf{Policy decision preserved} \\
\midrule
$P_t$ & Candidate pool after compression and deduplication & Which candidates to inspect, read, or curate \\
$C_t,I_t$ & Curated output set and importance tags, warm-started by auto-seeding & Which documents to add, remove, promote, or demote \\
$D_t$ & Full-text memory for retrieved documents & Which seen documents to revisit through \texttt{review\_docs} \\
$G_t$ & Evidence graph over entities, dates, and documents & Which bridge, singleton, or relation to pursue next \\
$V_t$ & Verification records for policy-written claims & Which claim to check and which documents to test \\
$H_t$ & Search history and result summaries & When to diversify, backtrack, or continue a search thread \\
$B_t$ & Budget-safe renderer and context marker & When to search, read, summarize, or stop \\
\bottomrule
\end{tabular}
}
\end{table}

\wm has two tiers. The prompt-facing tier renders compact state:
$P_t$, $C_t$, $I_t$, $G_t$, $V_t$, $H_t$, and $B_t$. The outer tier $D_t$
stores full text and metadata for retrieved chunks, which the policy can revisit
through \texttt{review\_docs} or \texttt{read\_document}. Thus the prompt
carries actionable search state rather than the full retrieval transcript.

Full state-transition algorithms and trajectory-level examples are given in
Appendices~\ref{app:harness-algorithms} and~\ref{app:case-studies}.

\subsection{Policy actions as edits over working memory}
\label{sec:method:action}

\sys exposes five classes of actions: retrieval, memory inspection, curation,
verification, and termination. Retrieval actions bring new evidence into the
episode. \texttt{fan\_out\_search} runs up to five diverse queries in parallel,
\texttt{search\_corpus} performs targeted hybrid search, \texttt{grep\_corpus}
performs exact pattern matching, and \texttt{read\_document} returns full text
for a known document ID. Their outputs are not simply appended to the prompt.
After compression and deduplication, they update the candidate pool $P_t$ and
the full-text store $D_t$.

\textbf{Curation.}
The central state-editing action is \curate, which updates the final output set
$C_t$ and its importance map $I_t$. The action specifies documents to add,
documents to remove, and importance tags for added documents. Tags take one of
four values: \texttt{very\_high}, \texttt{high}, \texttt{fair}, or
\texttt{low}. The curated set is capped at $M{=}30$ documents; when the cap is
reached, the harness evicts the lowest-importance documents first
(Appendix~\ref{app:importance}). This gives the policy an explicit language for
confidence and priority, while the harness handles the mechanical capacity
constraint.

\textbf{Auto-seeding.}
A blank curated set provides little learning signal on hard queries, since many
early rollouts terminate with the same empty output. To avoid this, the first
successful search automatically seeds $C_t$ with the top $k{=}8$ reranked
results from $P_t$, each assigned $I_t(d)=\texttt{fair}$. The policy must then
refine this tentative set by promoting strong documents and removing weak ones.
Auto-seeding does not decide relevance for the model. It changes the initial
state from construction from scratch to refinement, so that early trajectories
differ in how they edit candidates. The BC+ and SEC trajectories in
Appendices~\ref{app:case-bc643} and~\ref{app:case-sec497} show this pattern:
the first search seeds plausible candidates, and later \curate calls promote
documents that survive reading, exact search, or verification.

\textbf{Verification and memory review.}
\verify lets the policy write a claim and select remembered documents to test;
the harness checks support against text in $D_t$ and records yes/no judgments
with rationales in $V_t$. Thus the policy decides what requires verification,
while the harness performs the mechanical support check. \texttt{review\_docs}
re-renders documents already stored in $D_t$, allowing the policy to revisit
evidence before promoting, demoting, removing, or verifying it without issuing
another corpus search.


\subsection{Derived-state rendering: evidence graph, compression, and deduplication}
\label{sec:method:render}

The harness does not only store retrieved text. It derives compact signals from
the search history and renders them into each observation, so the policy sees a
structured search state rather than a raw accumulation of tool outputs.

\textbf{Evidence graph.}
The evidence graph $G_t$ summarizes cross-document structure. Each chunk entering
$P_t$ is scanned with a lightweight regex extractor for three entity classes:
multi-word capitalized proper nouns, four-digit years or decades, and numeric
dates. The harness maintains two maps, \texttt{entity\_to\_docs} and
\texttt{doc\_to\_entities}, and renders a compact block with frequent entities,
bridge documents, and singleton entities. Frequent entities are the top entities
by document count. Bridge documents contain two or more frequent entities and
are natural candidates for \verify or high-importance promotion. Singleton
entities appear in only one document and often provide leads for follow-up
search.

\begin{wrapfigure}{r}{0.50\textwidth}
\vspace{-1.8em}
\begin{tcolorbox}[
  width=0.5\textwidth,
  colback=white,
  colframe=black!85,
  boxrule=0.8pt,
  arc=4pt,
  left=4pt,
  right=4pt,
  top=4pt,
  bottom=4pt,
  boxsep=0pt
]
\small\ttfamily
[Evidence Graph] Entities appearing in multiple docs (bridges):\\
\hspace*{1em}Brussels: \textcolor{docred}{22816\_0}, \textcolor{docblue}{91442\_3}, 62390\_5\\
\hspace*{1em}1878: \textcolor{docred}{22816\_0}, 11293\_2\\
\hspace*{1em}Grande Synagogue: \textcolor{docred}{22816\_0}, \textcolor{docblue}{91442\_3}\\
\hspace*{1em}(2 singleton entities, potential hops)
\end{tcolorbox}
\vspace{-1.2em}
\end{wrapfigure}

The example illustrates how the graph makes evidence overlap visible. Document
\textcolor{docred}{22816\_0} connects all three rendered entities, while
\textcolor{docblue}{91442\_3} connects two of them. The policy can therefore
treat these documents as candidate bridges without rereading the full
transcript. More generally, the rendering changes questions such as ``what have
I seen about entity $X$?'' from a full-context reread into a direct lookup in
working memory. The harness builds the graph with one regex pass per chunk and a
top-$k$ scan at render time; the policy decides which bridge to verify, which
singleton to pursue, and which relation matters for the query.

\textbf{Result summaries and budget markers.}
After each retrieval or memory action, the harness records the tool used, the
number of returned and novel documents, updates to $P_t$ and $C_t$, and the
current context budget $B_t$. Thus the policy observes search progress and
remaining budget as explicit state.
\
\textbf{Sentence-BM25 compression.}
To prevent search observations from dominating the context, the harness
compresses outputs from \texttt{search\_corpus}, \texttt{fan\_out\_search}, and
\texttt{grep\_corpus} by scoring sentences with BM25 against the query and
rendering the top $K{=}4$ sentences in original order. Explicit
\texttt{read\_document} calls still return full text from $D_t$.
\
\textbf{Two-level deduplication.}
\sys removes repeated evidence by both chunk ID and content fingerprint.
Near-duplicates are omitted from the rendered state, while reward accounting
still credits all relevant evidence encountered during the trajectory.
\
The same budgeted context builder is used for teacher generation, SFT replay,
RL rollout, and evaluation. It combines the system prompt, rendered working
memory, recent actions and observations, and result summaries, while truncating
older reasoning and clipping long observations as needed.

\subsection{Training: SFT and RL}
\label{sec:training}

Training follows the same division of labor as the harness: SFT teaches the
model to operate the interface, while RL improves search decisions over the
state maintained by that interface.\footnote{We use Tinker
(\url{https://thinkingmachines.ai/tinker/}) for model training.}

\textbf{SFT for harness operation.}
The role of SFT is deliberately narrow. Since the harness carries much of the
bookkeeping load, the supervised prior only needs to teach tool-call format,
the search$\to$curate rhythm, importance-tagging discipline, memory review, and
the \verify-before-promote norm. A single teacher, GPT-5.4, runs as a live agent
inside the full \sys harness, using the same tools and working-memory rendering
later shown to the student. Turn-level guidance encourages curation after
search, verification before high-confidence promotion, backtracking after
low-yield searches, and early termination once the curated set is sufficient.
\
We keep trajectories that are format-valid, return at least one document, and
achieve final curated recall $\geq 0.10$. Each retained trajectory is replayed
through the same \wm and context builder used by RL, then expanded into one
supervised datum per turn. SFT runs on \texttt{gpt-oss-20b} via LoRA
with rank $32$ for $3$ epochs; the step-$550$ checkpoint initializes RL.

\textbf{RL over full search episodes.}
Starting from the SFT checkpoint, we train with on-policy CISPO
\citep{cispo2025,scalerl2025} and within-group advantage normalization on SEC
training queries. RL uses full-trajectory rollouts, terminal-only reward, a
$40$-turn cap, and no KL anchor. Groups whose rollouts receive identical
rewards are dropped from the gradient. Full rollout scale and optimization
hyperparameters are in Appendix~\ref{app:reward}.

\textbf{Reward.}
The terminal reward combines search-quality signals, answer-evidence shaping, a
tool-diversity bonus, and a turn penalty. Episodes that terminate with an empty
curated set short-circuit to $\pi_\emptyset=-0.2$; otherwise,
\begin{equation*}
\begin{aligned}
\mathcal{R} \;=\;& \underbrace{w_F\,F_\beta}_{\text{set quality}}
       + \underbrace{w_\tau\,\rho_\tau}_{\text{trajectory coverage}}
       + \underbrace{w_{\mathrm{A}}\,\rho_{\mathrm{A}}
                    + w_{\tau\mathrm{A}}\,\rho_{\tau\mathrm{A}}}_{\text{answer evidence}}
       + \underbrace{B_{\mathrm{A}}\,\mathbf{1}[\rho_{\mathrm{A}}>0]}_{\text{answer bonus}} \\
       & + \underbrace{w_{\mathrm{div}}\,\min(\nu/\nu_0,\,1)}_{\text{tool diversity}}
       - \underbrace{w_{\mathrm{miss}}\,(\rho_{\tau\mathrm{A}}-\rho_{\mathrm{A}})_+}_{\text{answer miss}}
       - \underbrace{\pi_{\mathrm{turn}}(t)}_{\text{turn penalty}} .
\end{aligned}
\end{equation*}
For any successfully formatted episode, the reward is clipped below by
$\mathcal{R}\geq 10^{-3}$. Here $\rho_\tau$ is trajectory recall,
$\rho_{\mathrm{A}}$ is curated final-answer recall,
$\rho_{\tau\mathrm{A}}$ is trajectory final-answer recall, $\nu$ is the number
of distinct tools used, and $\nu_0$ is the diversity target. $F_\beta$ uses
$\beta{=}2$, weighting recall four times more than precision. 
The reward separates discovery from selection: trajectory terms
credit evidence found anywhere in the pool, curated terms credit evidence
selected for the final output, and the answer-miss penalty penalizes
trajectories that found answer evidence but failed to promote it. Full
hyperparameters and code variables are listed in Appendix~\ref{app:reward}.


\begin{table}[t]
\centering
\small
\caption{\textbf{Search quality across benchmarks.}
We report curated-set Recall, Final-Answer Recall, and Trajectory Recall.
Green and blue rows denote the overall best models among open-source small models and
frontier models, respectively. Results are averaged over three runs.}
\label{tab:search_quality}
\resizebox{\textwidth}{!}{
\begin{tabular}{llcccccccc}
\toprule
\textbf{Model} & \textbf{Metric} & \textbf{BC+} & \textbf{Web} & \textbf{Patents} & \textbf{SEC}
               & \textbf{LongSeal} & \textbf{Seal0} & \textbf{FRAMES} & \textbf{HotpotQA} \\
\midrule
\multicolumn{10}{l}{\textbf{Open-source small models}} \\
\midrule
\rowcolor{bestcolor}
& Recall               & 0.584 & 0.787 & 0.890 & 0.589 & 0.891 & 0.551 & 0.710 & 0.841 \\
\rowcolor{bestcolor}
& Final-Answer Recall  & 0.667 & 0.845 & 0.890 & 0.659 & 0.891 & 0.551 & 0.710 & 0.841 \\
\rowcolor{bestcolor}
\multirow{-3}{*}{\sys (20B)}
& Trajectory Recall    & 0.665 & 0.881 & 0.973 & 0.677 & 0.944 & 0.708 & 0.749 & 0.860 \\
\midrule
\multirow{3}{*}{Context-1 (20B)}
& Recall               & 0.500 & 0.689 & 0.857 & 0.467 & 0.563 & 0.367 & 0.638 & 0.746 \\
& Final-Answer Recall  & 0.618 & 0.775 & 0.857 & 0.498 & 0.563 & 0.367 & 0.638 & 0.746 \\
& Trajectory Recall    & 0.567 & 0.823 & 0.975 & 0.507 & 0.858 & 0.697 & 0.787 & 0.835 \\
\midrule
\multirow{3}{*}{GPT-OSS-20B}
& Recall               & 0.109 & 0.174 & 0.393 & 0.133 & 0.253 & 0.250 & 0.391 & 0.395 \\
& Final-Answer Recall  & 0.156 & 0.252 & 0.393 & 0.166 & 0.253 & 0.250 & 0.391 & 0.395 \\
& Trajectory Recall    & 0.246 & 0.512 & 0.742 & 0.252 & 0.816 & 0.632 & 0.702 & 0.819 \\
\midrule
\multirow{3}{*}{{Tongyi DeepResearch 30B}} 
& Recall               & 0.381 & 0.607 & 0.780 & 0.419 & 0.855 & 0.577 & 0.561 & 0.751 \\
& Final-Answer Recall  & 0.428 & 0.679 & 0.780 & 0.433 & 0.855 & 0.577 & 0.561 & 0.751 \\
& Trajectory Recall    & 0.413 & 0.722 & 0.780 & 0.478 & 0.855 & 0.631 & 0.654& 0.852 \\
\midrule
\multirow{3}{*}{Search-R1 (32B)}
& Recall               & 0.053 & 0.135 & 0.405 & 0.109 & 0.514 & 0.312 & 0.329 & 0.454 \\
& Final-Answer Recall  & 0.074 & 0.240 & 0.405 & 0.155 & 0.514 & 0.312 & 0.329 & 0.454 \\
& Trajectory Recall    & 0.053 & 0.135 & 0.405 & 0.109 & 0.514 & 0.312 & 0.329 & 0.454 \\
\midrule
\multirow{3}{*}{Qwen3 (32B)}
& Recall               & 0.072 & 0.132 & 0.257 & 0.058 & 0.427 & 0.279 & 0.164 & 0.342 \\
& Final-Answer Recall  & 0.092 & 0.211 & 0.257 & 0.072 & 0.427 & 0.279 & 0.164 & 0.342 \\
& Trajectory Recall    & 0.154 & 0.410 & 0.577 & 0.215 & 0.704 & 0.568 & 0.402 & 0.535 \\
\midrule
\multicolumn{10}{l}{\textbf{Frontier models}} \\
\midrule
\rowcolor{lightblue}
& Recall               & 0.619 & 0.819 & 0.915 & 0.589 & 0.823 & 0.636 & 0.814 & 0.893 \\
\rowcolor{lightblue}
& Final-Answer Recall  & 0.736 & 0.928 & 0.915 & 0.696 & 0.823 & 0.636 & 0.814 & 0.893 \\
\rowcolor{lightblue}
\multirow{-3}{*}{Opus-4.6}
& Trajectory Recall    & 0.634 & 0.832 & 0.939 & 0.601 & 0.876 & 0.682 & 0.878 & 0.910 \\
\midrule
\multirow{3}{*}{GPT-5.4}
& Recall               & 0.443 & 0.800 & 0.909 & 0.452 & 0.824 & 0.555 & 0.815 & 0.870 \\
& Final-Answer Recall  & 0.499 & 0.858 & 0.909 & 0.465 & 0.824 & 0.555 & 0.815 & 0.870 \\
& Trajectory Recall    & 0.476 & 0.821 & 0.919 & 0.464 & 0.885 & 0.691 & 0.865 & 0.898 \\
\midrule
\multirow{3}{*}{Sonnet-4.6}
& Recall               & 0.505 & 0.744 & 0.832 & 0.411 & 0.783 & 0.577 & 0.787 & 0.863 \\
& Final-Answer Recall  & 0.597 & 0.892 & 0.832 & 0.480 & 0.783 & 0.577 & 0.787 & 0.863 \\
& Trajectory Recall    & 0.535 & 0.767 & 0.845 & 0.421 & 0.840 & 0.658 & 0.859 & 0.878 \\
\midrule
\multirow{3}{*}{Kimi-K2.5}
& Recall               & 0.593 & 0.707 & 0.833 & 0.454 & 0.661 & 0.455 & 0.697 & 0.776 \\
& Final-Answer Recall  & 0.670 & 0.843 & 0.833 & 0.478 & 0.661 & 0.455 & 0.697 & 0.776 \\
& Trajectory Recall    & 0.739 & 0.878 & 0.945 & 0.549 & 0.911 & 0.655 & 0.840 & 0.837 \\
\midrule
\multirow{3}{*}{GPT-OSS-120B}
& Recall               & 0.508 & 0.478 & 0.570 & 0.395 & 0.464 & 0.400 & 0.495 & 0.657 \\
& Final-Answer Recall  & 0.641 & 0.664 & 0.570 & 0.504 & 0.464 & 0.400 & 0.495 & 0.657 \\
& Trajectory Recall    & 0.682 & 0.796 & 0.824 & 0.488 & 0.914 & 0.689 & 0.863 & 0.897 \\
\bottomrule
\end{tabular}
}
\end{table}

\section{Experiments}
\subsection{Experimental Setup}
\label{sec:setup}

\textbf{Benchmarks \& Data.} We evaluate on eight retrieval benchmarks:
BrowseComp+~\cite{chen2025browsecompplus}, \emph{Web} synthetic,
\emph{Patents} from USPTO office actions, \emph{SEC} filings~\citep{bashir2026context1},
LongSealQA and Seal0QA~\cite{pham2025sealqa},
FRAMES~\cite{krishna2024frames}, and HotpotQA~\cite{yang2018hotpotqa}.
They span web, finance, patents, and multi-hop QA, and use both
Chroma-backed corpora and a shared Serper$+$Jina web backend.
\sys's SFT data spans four benchmark families (BC+, Web, Patents,
SEC), while the RL stage optimizes only on SEC data. We therefore
separate the four source-family benchmarks (BC+, Web, Patents, SEC)
from four held-out transfer benchmarks not used for \sys SFT or RL
(LongSealQA, Seal0QA, FRAMES, HotpotQA). Per-benchmark statistics are
in Appendix~\ref{app:benchstats}. The raw SFT quotas are BC+/300,
SEC/250, Patents/150, Web/150, Web-simple/75, and SEC-simple/75; after
filtering, $899$ trajectories remain. We use the training split of SEC\footnote{\url{https://github.com/chroma-core/context-1-data-gen/blob/main/agentic_search_data_gen/}} (3,453 queries in total) for RL training.

\textbf{Search-quality metrics.}
We evaluate search quality by evidence coverage. \emph{Recall} measures
coverage of all annotated relevant documents in the final curated set.
\emph{Trajectory Recall} measures the same coverage over every document
encountered during the episode, before final selection. \emph{Final-Answer
Recall} measures coverage of the gold answer documents only. Formal
definitions, including the distinction between answer documents and supporting
documents, are given in Appendix~\ref{app:metric-defs}. We focus on
recall-oriented metrics since the search agent is designed to collect evidence
needed for answering. Precision and $F_1$ are secondary because curated-set size
is part of the agent interface.

\textbf{Evaluation protocol and baselines.}
We evaluate on the benchmark splits summarized in
Table~\ref{tab:bench-stats}. All methods use the same retrieval primitives,
web backend, Qwen3-Reranker-8B reranker, and final budget of at most
$30$ returned documents, following Context-1~\citep{bashir2026context1}.
\sys runs at temperature $1.0$ with a $40$-turn cap. Context-1 and proprietary
frontier LLMs (GPT-5.4, Sonnet-4.6, Opus-4.6, Kimi-K2.5), used as zero-shot
retrievers, run under the Context-1 harness at temperature $1.0$ with a
$64$-turn cap; Appendix~\ref{app:harness-confound} quantifies the harness
contribution alone (a $+4$-point recall lift on the same LLM).
Open-model baselines include base \texttt{gpt-oss-20b}/\texttt{gpt-oss-120b},
Qwen3-32B, Context-1, Search-R1~32B~\citep{jin2025searchr1}, and Tongyi
DeepResearch 30B~\citep{team2025tongyi}. For Search-R1 and Tongyi
DeepResearch, we use their released agent harnesses while standardizing the
external search and reranking APIs. Because these harnesses do not natively
produce a curated set capped at $30$ documents, we apply reranking to obtain the final top-$30$ documents; details are given in
Appendix~\ref{app:eval-recipe}. Full evaluation details are in
Appendices~\ref{app:metric-defs} and~\ref{app:benchstats}.


\subsection{Results}
\label{sec:results}

Table~\ref{tab:search_quality} reports Recall, Final-Answer Recall,
and Trajectory Recall on all eight benchmarks. \sys is the strongest
open retrieval subagent we evaluate, reaching $0.730$ average curated
recall and improving over the next open subagent, Tongyi DeepResearch
30B, by $+11.4$ points.
It is also competitive with much larger frontier retrievers: it has
higher average curated recall than GPT-5.4, Sonnet-4.6, Kimi-K2.5,
and GPT-OSS-120B under this protocol, with Opus-4.6 the only frontier
retriever ahead on average.

\begin{wrapfigure}{r}{0.5\textwidth}
\vspace{-0.8em}
\centering
\includegraphics[width=\linewidth]{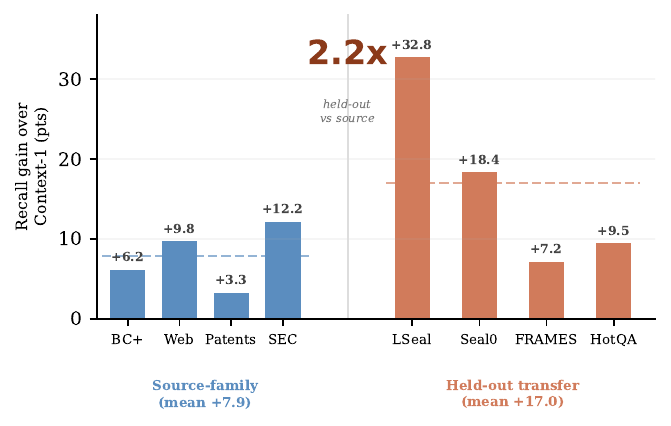}
\vspace{-1.5em}
\caption{\textbf{Transfer pattern.} \sys gains more on held-out
transfer benchmarks ($+17.0$ pts mean) than on source-family
benchmarks ($+7.9$ pts), a $2.2\times$ gap over Context-1.}
\label{fig:ood-asymmetry}
\end{wrapfigure}

\paragraph{Transfer is the clearest signature of the mechanism.}
Absolute recall is useful, but the structure of the gains is more
diagnostic. \sys is not trained from a single dataset: its SFT stage
uses BC+, Web, Patents, and SEC trajectories. The stricter transfer
test is therefore the four benchmarks excluded from both \sys SFT and
RL. If the gains came mainly from replaying source-family behaviors,
the advantage should shrink on these held-out tasks. We observe the
opposite.

A standard machine-learning prior would expect gains to be strongest
on benchmarks closest to the SFT and RL data. Across the four
source-family benchmarks (BC+, Web, Patents, SEC), \sys gains $+6.2$,
$+9.8$, $+3.3$, and $+12.2$ recall points over the closest open
baseline (mean $+7.9$).

Across the four
\emph{held-out transfer benchmarks} excluded from \sys SFT and RL
(LongSealQA, Seal0QA, FRAMES, HotpotQA), \sys gains
$+32.8$, $+18.4$, $+7.2$, and $+9.5$ points
(\emph{mean $+17.0$}). This is a $\boldsymbol{2.2\times}$ larger gain
on the benchmarks furthest from any training data
(Figure~\ref{fig:ood-asymmetry}). The mechanism is straightforward:
the policy is learning operations over a domain-general search state:
refine the auto-seeded set, read bridge entities from the evidence
graph, re-inspect uncertain candidates, verify before promotion, and
submit a compact curated set. Those operations transfer more naturally
than domain-specific patterns stored in weights. The FRAMES and
HotpotQA trajectories in Appendices~\ref{app:case-frames461}
and~\ref{app:case-hotpot-snl} show the same operations outside the
SFT/RL families: fan-out or targeted search proposes candidate bridges,
then reading, exact matching, and curation resolve the final evidence.

\begin{wrapfigure}{r}{0.4\textwidth}
\vspace{-2em}
\centering
\includegraphics[width=\linewidth]{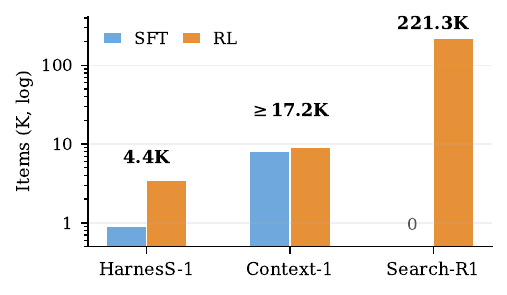}
\caption{\textbf{Training-data scale.} Paired bars show reported
unique task/query units used for SFT and RL. 
}
\label{fig:training-data-scale}
\vspace{-1.5em}
\end{wrapfigure}

\paragraph{Data scale.}
The transfer pattern is not explained by simply using more training
data. \sys uses $899$ filtered SFT trajectories and RL on the SEC
train split ($3{,}453$ queries), for $4{,}352$ unique training items.
Context-1 reports over $8$K synthetic SFT tasks and its RL trainer uses
$9{,}159$ unique queries (the four full train splits sum to $8{,}859$,
plus two capped 150-query curriculum splits). Search-R1 uses no SFT
warmup and trains RL on the merged NQ+HotpotQA set ($221{,}328$ rows).
Figure~\ref{fig:training-data-scale} therefore compares unique
task/query counts, not rollout count or token compute. The point is
not that less data is always sufficient, but that \sys moves much of
the behavioral prior into the stateful interface, allowing small SFT
and focused RL to transfer.

\begin{table}[t]
\centering
\small
\caption{\textbf{Inference-time component ablation on
BrowseComp+} (same trained \sys
checkpoint). Each row disables exactly one harness mechanism.
$\Delta$Recall (\%) and $\Delta$FA (\%) report the
\emph{relative} change versus full \sys, i.e.\ $100\times
(\text{ablated}-\text{full})/\text{full}$. The rightmost column
points to the per-mechanism failure mode with a representative
example trajectory.}
\label{tab:ablation}
\resizebox{\textwidth}{!}{
\begin{tabular}{lccccc}
\toprule
\textbf{Configuration} & \textbf{Recall} & \textbf{$\Delta$Recall (\%)} & \textbf{FA Recall} & \textbf{$\Delta$FA (\%)} & \textbf{Failure mode} \\
\midrule
\rowcolor{bestcolor}
Full \sys (all mechanisms enabled)                  & 0.584 & ---      & 0.667 & ---      & --- \\
\midrule
$-$ Importance tags (binary \curate, FIFO eviction) & 0.560          & $-4.1$  & 0.614          & $-7.9$  & §\ref{app:err:importance-tags} \\
$-$ Sentence-BM25 compression (raw chunks)          & 0.585          & $+0.2$  & 0.620          & $-7.0$  & §\ref{app:err:sentence-compress} \\
$-$ Auto-seed on first search                       & 0.582          & $-0.3$  & 0.624          & $-6.4$  & §\ref{app:err:auto-seed} \\
$-$ Evidence graph hidden in observations           & 0.569          & $-2.6$  & 0.631          & $-5.4$  & §\ref{app:err:evidence-graph} \\
$-$ \verify returns ``unavailable''                 & 0.566          & $-3.1$  & 0.641          & $-3.9$  & §\ref{app:err:verify} \\
$-$ \texttt{review\_docs} returns ``unavailable''   & 0.598          & $+2.4$  & 0.641          & $-3.9$  & §\ref{app:err:review-docs} \\
$-$ Content-fingerprint dedup (chunk-ID dedup kept) & 0.611          & $+4.6$  & 0.678          & $+1.6$  & §\ref{app:err:content-dedup} \\
\midrule
All \sys harness mechanisms disabled                & 0.513          & $-12.2$ & 0.624          & $-6.4$  & §\ref{app:err:all-disabled} \\
\bottomrule
\end{tabular}}
\end{table}

\paragraph{Inference-Time Component Ablation}
\label{sec:ablation}
We run the trained \sys checkpoint on $100$ paired BrowseComp+
test queries with individual harness mechanisms disabled at
inference time (no retraining). Each row of
Table~\ref{tab:ablation} disables exactly one mechanism;
\verify and \texttt{review\_docs} are switched off by returning
``unavailable.'' The bottom row disables \emph{all} \sys harness
mechanisms simultaneously while keeping the trained policy
unchanged.

\textbf{Six of seven mechanisms produce clean FA-recall drops}
($-3.9\%$ to $-7.9\%$ relative) with a consistent behavioural
signature: on queries the full system solves but the ablated
condition fails, \texttt{search\_corpus} rises by $3$--$7$
points and \texttt{read\_document}/\verify drop $2$--$6\times$.
Removing a state mechanism does not just lose information ---
the trained policy reverts to a wide, shallow, search-dominated
mode and never converges on the relevant documents. The four
largest FA drops correspond to the mechanisms with coupled
training roles: importance tags
(§\ref{app:err:importance-tags}), sentence compression
(§\ref{app:err:sentence-compress}), auto-seed
(§\ref{app:err:auto-seed}), and the evidence graph
(§\ref{app:err:evidence-graph}).
\
\textbf{Content dedup is the only ablation that nominally
helps} ($+4.6\%$ Recall, $+1.6\%$ FA, both relative). BC+ qrels
occasionally contain near-duplicate gold documents, which our
MinHash--LSH dedup ($\theta_{\text{Jaccard}}{=}0.85$) sometimes
collapses, costing one gold ID per affected query. Dedup is a
token-budget mechanism, not a recall mechanism: it shrinks the
curated context for downstream answering with negligible loss
in what the agent actually retrieves. We report it rather than
hide it; the trade-off is the design intent
(§\ref{app:err:content-dedup}).
\
\textbf{Mechanisms compose.} Disabling \emph{all} harness
mechanisms at once gives Recall $0.513$ ($-12.2\%$) and FA
$0.624$ ($-6.4\%$) --- a larger relative Recall drop than any
single ablation. The trained policy keeps searching but cannot
prioritize what it sees: the harness is where per-turn search
bandwidth is converted into a discriminative curated set, and
removing it leaves exploration intact but the decision
substrate gone (§\ref{app:err:all-disabled}).

\begin{wrapfigure}{r}{0.32\textwidth}
\vspace{-0.5em}
\centering
\includegraphics[width=\linewidth]{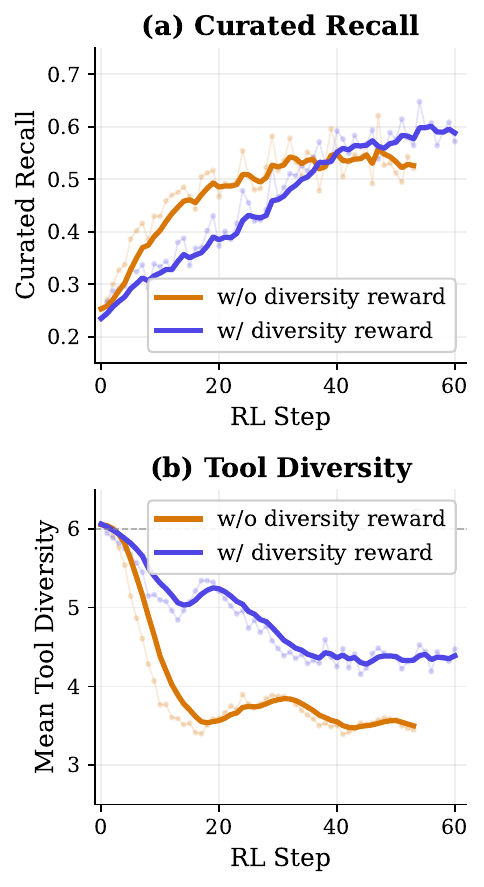}
\caption{\textbf{Training dynamics.} \textbf{(a)}~Recall
during RL training. \textbf{(b)}~Mean tool diversity. \textcolor{divoff}{Without the
diversity bonus}, recall plateaus and
tool use collapses; \textcolor{divon}{with the bonus},
diversity stabilises and final recall is higher.}
\label{fig:dynamics}
\vspace{-3em}
\end{wrapfigure}


\paragraph{Discovery and selection.}
Trajectory recall shows that \sys usually discovers the relevant
evidence: across domains, the harness surfaces a large pool of gold
documents before final curation. The remaining gap to Opus-4.6 on BC+
final-answer recall ($0.658$ vs.\ $0.736$), despite similar trajectory
recall, is therefore mainly a selection gap rather than a discovery gap.
Qualitative examples in Appendices~\ref{app:case-bc643}
and~\ref{app:case-sec497} show cases where the right document is seen
but not preserved in the final curated set.

\paragraph{Training Dynamics}
\label{sec:dynamics}
Figure~\ref{fig:dynamics} compares two RL runs that differ only in
whether the tool-diversity reward $w_{\mathrm{div}}$ is active.
Both start from the same SFT checkpoint and use the same
hyperparameters otherwise. The comparison shows that a rich harness
alone is insufficient: without RL-compatible shaping, the policy
collapses to a narrow search-heavy strategy.

Without $w_{\mathrm{div}}$, the agent quickly learns to issue many
\texttt{fan\_out\_search} calls, which improves trajectory recall but
bypasses curation and verification. Tool diversity falls from
${\sim}6$ to ${\sim}3.5$, and curated recall plateaus at ${\sim}0.53$.
The agent finds relevant documents, but does not reliably promote the
right ones into the final output set.

With $w_{\mathrm{div}}$ active, tool use remains broader: diversity
stabilises around ${\sim}4.30$, and the agent regularly combines search
with curation and verification. Early recall improves more slowly
because the policy spends more turns editing the candidate set, but
final curated recall is higher, reaching ${\sim}0.60$.
\
This demonstrates the third RL-compatibility requirement. The harness
must not only expose useful tools; the reward must also make those
tools learnable and worth using. Otherwise, RL exploits the easiest
reward path and collapses back toward a minimal search-only harness.

\paragraph{Modular RAG answer accuracy.}
The search gains also translate to downstream answer quality. When each
subagent's curated set is passed to frozen frontier generators, better
curated sets yield higher answer accuracy, and \sys outperforms the
open subagents in the reported modular RAG setting. Full plots and
protocol details are in Appendix~\ref{sec:answeracc}.




\section{Conclusion}
\label{sec:conclusion}

We introduced \sys, a 20B search agent trained with reinforcement learning
inside a stateful harness. The key idea is to separate semantic search decisions
from recoverable bookkeeping: the policy decides what to search, curate, verify,
and submit, while the harness maintains candidate pools, importance-tagged
evidence, verification records, evidence links, and budget-aware context
rendering.

Across eight benchmarks, \sys achieves the strongest average recall among the
open search agents we evaluate and remains competitive with much larger
frontier-model searchers. Its gains on held-out transfer benchmarks and
component ablations suggest that the harness is not merely an implementation
detail, but a central part of what the policy learns to use.

These results point to stateful harness design as an important direction for
retrieval-agent RL. Future work can make the evidence graph more intelligent,
for example by replacing regex-based extraction with learned entity linking,
relation extraction, and uncertainty-aware evidence organization. We discuss
connections to agentic search, tool orchestration, and harness
engineering in Appendix~\ref{app:related}, and limitations, ethics, and broader
impact in Appendix~\ref{sec:limitations-ethics-impact}.
\bibliographystyle{plain}
\bibliography{references}

\clearpage
\DoToC

\appendix

\section{Related Work}
\label{app:related}

\paragraph{Harness engineering.}
Harness engineering designs the environment layer between a language
model and its task~\citep{openai2026harness,martin2026managed}. Recent
work shows that the same model can vary by tens of points when placed
in different harnesses~\citep{yang2024sweagent,
xia2024agentlessdemystifyingllmbasedsoftware}, making
the interface a major determinant of system-level performance. Tool
orchestration~\citep{qin2023toolllm,jin2025searchr1,liu2025webexplorer,su2025toolorchestraelevatingintelligenceefficient}
usually fixes this interface and trains the policy to call the exposed
tools more effectively. Harness engineering asks a lower-level design
question: what interface should the policy operate?

Most existing harnesses are designed by hand~\citep{yang2024sweagent,
zhang2025generalmodularharnessllm}, often as empirical responses to
model behaviors that may change as models improve~\citep{martin2026managed}.
Recent systems automate harness construction by searching over code
space~\citep{lou2026autoharness,lee2026meta} or configuration
space~\citep{sengupta2026harborautomatedharnessoptimization}. In
\sys, we study a complementary question: how should a retrieval harness
be shaped so that a trainable policy can use it well? Our answer is to
offload mechanical search bookkeeping into environment-side state, then
train the policy to make semantic decisions over that state.

\begin{keybox}
\textbf{Tool orchestration vs.\ stateful harnessing.}
Let an agent interface be
$\mathcal{I}=(\mathcal{A},\mathcal{O},T,r)$: actions,
observation renderer, environment transition, and reward. Tool
orchestration usually fixes an interface $\mathcal{I}_0$ and learns a
policy inside it:
\[
\max_\theta \; \mathbb{E}\!\left[R(\pi_\theta;\mathcal{I}_0)\right].
\]
Stateful harnessing changes $\mathcal{I}$ by adding the editable
retrieval state from Table~\ref{tab:state-offload}:
\[
s_t=(P_t,C_t,I_t,D_t,G_t,V_t,H_t,B_t).
\]
The transition is
\[
(s_t,a_t)\mapsto(s_{t+1},o_{t+1}),
\]
not merely $a_t\mapsto o_{t+1}$. The policy still learns tool use, but
it learns over a state representation maintained by the harness:
\[
\max_\theta \; \mathbb{E}\!\left[R(\pi_\theta;\mathcal{I}_{\sys})\right].
\]
\end{keybox}



\paragraph{Agentic search.}
Agentic search refers to large language  model agents that gather evidence through multiple rounds of query generation, reading returned documents, and deciding when to stop~\citep{zhang2025web}. 
Early work in this direction interleaved retrieval with reasoning in tight loops~\citep{li2025search, gao2025beyond, liu2025webexplorer, cognition2025swegrep}. ReAct~\citep{yao2023react}, Self-Ask~\citep{press2023measuring}, and IRCoT~\citep{trivedi2023interleaving} established the pattern of alternating reasoning and tool calls. However, prompting-based reasoning alone is often inconsistent with surface behavior failing to reliably translate into the intended decision patterns~\citep{turpin2023language,lanham2023measuring, han2025personalityillusionrevealingdissociation, han2026steer2adapt}.

Reinforcement learning post-training~\citep{schulman2017proximalpolicyoptimizationalgorithms, hu2025reinforce++, yu2025dapo, Guo_2025} has emerged as a powerful tool for shaping LLM reasoning, and a growing line of work brings it to retrieval and agentic search~\citep{song2025r1, chen2025researchlearningreasonsearch, jiang2025deepretrieval, zheng-etal-2025-deepresearcher}. DeepRetrieval~\citep{jiang2025deepretrieval} shows that LLMs can be trained with RL to interact directly with real search engines and retrievers, improving retrieval through learned query and search behavior. Search-R1~\citep{jin2025searchr1} trains a single LLM to interleave reasoning and search calls under a downstream question-answering reward. s3~\citep{jiang2025s3} decouples the searcher from the generator, training only the search policy for modular optimization with frozen generators. These methods work well at short horizons but as tasks grow longer and more complex, like report generation~\citep{wu2025webdancer, li2025websailor}, the policy is asked to do too much at once: planning the next search\citep{chen2026apexsearcheraugmentingllmssearch, sun2025decouplesearch}, remembering which documents have been seen~\citep{zhou2025mem1, packer2024memgpt}, tracking which constraints remain uncovered~\citep{chen2025browsecompplus}, and managing an ever-growing context~\citep{hong2025context}. Under this load, the reward signal grows sparse relative to the work the policy is actually doing causing the degradation of RL training~\citep{deng2025atom, wang2025spa}. Recent work begins to address parts of this overload. Context-1~\citep{bashir2026context1} trains a 20B retrieval agent that learns to selectively prune its own context during search, beginning to treat context management as part of the policy itself.

Across these systems, the harness of the agent remains a thin tool wrapper, and any state beyond 
the prune action must be reconstructed by the policy 
from a raw, append-only observation stream. 
\sys takes a different stance: instead of training a stronger 
policy on a thin interface, we make the interface itself part of the 
method. 
The harness is stateful, with environment-side working memory 
that absorbs the mechanical bookkeeping of multi-turn search. We frame this division of labor as 
cognitive offloading, and treat harness design itself as a central axis of agentic search.

\section{Limitations, Ethics, and Broader Impact}
\label{sec:limitations-ethics-impact}

\paragraph{Limitations.}
\sys is designed for evidence-seeking retrieval tasks where the goal is to
identify documents that support an answer. Most benchmarks in our evaluation are
needle-in-a-haystack or multi-hop search tasks with annotated relevant evidence.
This setting is important, but it does not cover all retrieval use cases.
Breadth-oriented research tasks, open-ended report generation, abstention under
missing evidence, and adversarial web environments remain outside the main scope
of this work.

Our harness also introduces several engineered components whose quality may vary
across domains. The evidence graph uses a lightweight regex-based extractor for
entities, dates, and years, rather than a full entity-linking system. The
\verify tool uses an LLM-based verifier as a proxy entailment model, which can
make mistakes on ambiguous, highly technical, or underspecified claims. The
sentence-level BM25 compressor may remove useful context when relevance depends
on discourse structure rather than local sentence overlap.

Evaluation is limited by benchmark size and annotation coverage. Some datasets
have non-negligible confidence intervals, and near-duplicate or incomplete qrels
can affect recall-oriented metrics. We therefore report multiple retrieval
metrics, trajectory-level diagnostics, ablations, and modular RAG answer
accuracy, but these measurements should still be interpreted as evidence of
retrieval behavior under the specified benchmark and harness conditions rather
than as a complete measure of real-world search reliability.

\paragraph{Ethics.}
This work studies how to train language-model search agents to retrieve
supporting evidence more effectively. The primary ethical motivation is to make
retrieval-augmented systems more transparent and auditable: instead of relying
only on a model's parametric memory, the system returns an explicit curated set
of documents that can be inspected, verified, and used by downstream models or
human users.

The main ethical risks come from misuse and over-reliance. A stronger search
agent could retrieve sensitive, private, copyrighted, or otherwise inappropriate
information if connected to an unrestricted corpus. It could also amplify
misleading evidence if the retrieval backend contains low-quality, biased, or
adversarial documents. Our method should therefore be deployed only with
appropriate corpus access controls, logging, rate limits, privacy filters, and
human oversight for sensitive domains.

The system should not be treated as a source of truth by itself. \sys retrieves
and curates evidence; it does not guarantee that the evidence is complete,
unbiased, or correctly interpreted. In high-stakes domains such as medicine,
law, finance, or public policy, any downstream answer should be reviewed by
qualified experts and validated against authoritative sources.

\paragraph{Broader impact.}
Stateful retrieval agents can reduce the cost of evidence gathering and make
long-horizon search more accessible. This may benefit scientific literature
review, legal and financial document analysis, fact-checking, education, and
other workflows where users need to locate and organize evidence across many
documents. Because \sys separates retrieval from generation, its outputs can be
audited more directly than closed-book answers.

At the same time, lower-cost search agents can also increase the scale of
undesirable information-seeking behavior, including privacy-invasive search,
automated evidence cherry-picking, and targeted collection of sensitive
information. These risks are not unique to our method, but stronger and cheaper
retrieval agents can make them easier to execute. Responsible deployment should
therefore restrict corpus access, monitor usage, and pair retrieval with
verification and human review.

Our planned release is limited to the retrieval subagent, harness code,
data-generation pipeline, and evaluation recipe, with documentation of intended
use and dataset provenance. It is not a general autonomous web agent and is not
intended to replace expert judgment in high-stakes decision-making.

\section{Tool Signatures}
\label{app:tools}

This section lists the tools exposed to the \sys policy. The tools are grouped
around the main operations in the harness: retrieving new candidates, revisiting
stored evidence, editing the curated set, verifying claims, and ending the
episode. The policy chooses among these tools, while the harness executes the
calls and updates the working-memory state described in
Table~\ref{tab:state-offload}.

Table~\ref{tab:tools} gives the functional signature of each tool used in all
teacher, SFT, RL, and evaluation runs.

\begin{table}[ht]
\centering
\small
\caption{\sys tool signatures.}
\label{tab:tools}
\begin{tabular}{ll}
\toprule
\textbf{Tool} & \textbf{Effect} \\
\midrule
\texttt{fan\_out\_search(queries)} & Up to five parallel hybrid searches, RRF + rerank \\
\texttt{search\_corpus(query)} & Single hybrid search (BM25 + dense), RRF, rerank \\
\texttt{grep\_corpus(pattern)} & Regex match over the corpus \\
\texttt{read\_document(doc\_id)} & Full-text read, reranked and truncated to budget \\
\texttt{review\_docs(doc\_ids)} & Re-render seen docs from \wm (no corpus call) \\
\curate(add, remove, importance) & Edit curated set with four-level importance tag \\
\verify(doc\_ids, claim) & Per-document LLM entailment check against the claim \\
\texttt{end\_search(reasoning)} & Terminate episode, return curated set \\
\midrule
\multicolumn{2}{l}{\emph{For backward compatibility (registered, no-op):}} \\
\texttt{prune\_chunks(\ldots)} & Returns ``managed via working memory; no pruning needed.'' \\
\bottomrule
\end{tabular}
\end{table}

\section{Reward \& Training Hyperparameters}
\label{app:reward}

Table~\ref{tab:reward-hp} lists the reward weights and penalty
coefficients used in the deployed RL run, alongside their
configuration names.

\begin{table}[ht]
\centering
\small
\caption{Reward hyperparameters for the deployed \sys RL run.
\textbf{Note on the empty-set penalty:} when the curated set is
empty at episode termination, the reward short-circuits to
$\pi_\emptyset = -0.2$, bypassing the rest of the formula. For
all other terminations, the formula in §\ref{sec:training} applies and
is clipped below by $\mathcal{R} \geq 10^{-3}$.}
\label{tab:reward-hp}
\begin{tabular}{lll}
\toprule
\textbf{Symbol} & \textbf{Code variable} & \textbf{Value} \\
\midrule
$w_F$ (set quality) & \texttt{OUTCOME\_WEIGHT} & $0.7$ \\
$w_\tau$ (trajectory recall) & \texttt{TRAJECTORY\_RECALL\_WEIGHT} & $0.3$ \\
$\beta$ ($F_\beta$ exponent) & \texttt{RECALL\_BETA} & $2.0$ \\
$w_{\mathrm{A}}$ (FA recall) & \texttt{FINAL\_ANSWER\_RECALL\_WEIGHT} & $0.8$ \\
$w_{\tau\mathrm{A}}$ (trajectory FA recall) & \texttt{TRAJECTORY\_FA\_RECALL\_WEIGHT} & $0.4$ \\
$B_{\mathrm{A}}$ (binary FA bonus) & \texttt{FINAL\_ANSWER\_BONUS} & $1.0$ \\
$w_{\mathrm{miss}}$ (FA miss penalty) & \texttt{FA\_MISS\_PENALTY\_WEIGHT} & $0.35$ \\
$\pi_\emptyset$ (empty-set short-circuit) & \texttt{NO\_CURATE\_PENALTY} & $-0.2$ \\
Turn penalty start & \texttt{TURN\_PENALTY\_MIN\_TURNS} & $20$ \\
Turn penalty max & \texttt{TURN\_PENALTY\_MAX} & $0.02$ \\
Max turns hard cap & \texttt{MAX\_TURNS} & $40$ \\
Format-reward floor & \texttt{MIN\_FORMAT\_REWARD} & $10^{-3}$ \\
KL coefficient (anchor to SFT) & \texttt{KL\_PENALTY\_COEF} & $0.0$ \\
$w_{\mathrm{div}}$ (tool-diversity bonus weight) & \texttt{TOOL\_DIVERSITY\_BONUS} & $\mathbf{0.15}$ \\
$\nu_0$ (tool-diversity target) & \texttt{TOOL\_DIVERSITY\_TARGET} & $\mathbf{6}$ \\
\bottomrule
\end{tabular}
\end{table}

We also report the training configuration for the released \sys
checkpoint. We separate hyperparameters for SFT, RL, the search environment,
retrieval infrastructure, and evaluation. The goal is to make the reported
training recipe reproducible and to clarify which settings belong to the model
optimization stage versus the harness and retrieval backend.

SFT is used only to teach the model how to operate the stateful interface. RL is
then applied over full search episodes using the same harness state renderer and
tool set. Table~\ref{tab:hp-train} lists the exact settings used in the
deployed run.

\begin{table}[ht]
\centering
\small
\caption{SFT and RL hyperparameters for the deployed \sys
checkpoint.}
\label{tab:hp-train}
\begin{tabular}{lll}
\toprule
\textbf{Stage} & \textbf{Setting} & \textbf{Value} \\
\midrule
SFT & Base model                 & \texttt{openai/gpt-oss-20b} (MoE) \\
SFT & Adapter                    & LoRA, rank $32$ \\
SFT & Optimiser                  & AdamW (Tinker default) \\
SFT & Learning rate              & $5\times10^{-6}$ \\
SFT & Batch size                 & $128$ \\
SFT & Max sequence length        & $32{,}768$ \\
SFT & Epochs                     & $3$ \\
SFT & Min recall filter          & $0.10$ (final output recall) \\
SFT & Per-trajectory expansion   & one datum per turn \\
SFT & Selected checkpoint        & step $550$ (sampler weights) \\
\midrule
RL  & Algorithm                  & on-policy CISPO (clip $[0,5]$) \\
RL  & Adapter                    & LoRA, rank $32$ \\
RL  & Learning rate              & $1\times10^{-5}$ \\
RL  & Queries per step (BATCH)   & $128$ \\
RL  & Rollouts per query (GROUP) & $8$ \\
RL  & Total rollouts per step    & $1{,}024$ \\
RL  & Substeps per step          & $4$ \\
RL  & Total steps                & $80$ \\
RL  & Total rollouts             & ${\sim}82$K \\
RL  & Rollout temperature        & $1.0$ \\
RL  & Generation budget          & $2{,}048$ tokens \\
RL  & Model context limit        & $32{,}768$ tokens \\
RL  & KL anchor                  & disabled \\
RL  & Drop constant-reward groups& yes \\
RL  & Dataset                    & \texttt{sec} (single-domain) \\
\midrule
Env & MAX\_TURNS                 & $40$ \\
Env & RECENT\_K (recent window)  & $5$ \\
Env & SEARCH\_DISPLAY\_LIMIT     & $10$ \\
Env & MAX\_OBS\_CHARS            & $15{,}000$ \\
Env & SEARCH\_TOKEN\_BUDGET      & $4{,}096$ \\
Env & SENTENCE\_COMPRESS\_K      & $4$ \\
Env & MINHASH\_DEDUP\_THRESHOLD  & $0.85$ \\
Env & MINHASH\_NUM\_PERM         & $64$ \\
Env & EVIDENCE\_GRAPH\_MAX\_ENT  & $8$ \\
Env & AUTO\_POPULATE\_TOP\_K     & $8$ \\
Env & MAX\_CURATED\_DOCS         & $30$ \\
Env & DOC\_SNIPPET\_CHARS        & $120$ \\
\midrule
Retr & Retrieval                 & hybrid BM25 $+$ dense, RRF \\
Retr & Reranker                  & Qwen3-Reranker-8B (Baseten-hosted) \\
Retr & Reranker max tokens       & $4{,}096$ \\
\midrule
Eval & Sampling temperature      & $1.0$ \\
Eval & Max tokens per turn       & $2{,}048$ (\sys), $4{,}096$ (Context-1) \\
Eval & Max trajectory length     & $40$ turns (\sys), $64$ turns (Context-1) \\
\bottomrule
\end{tabular}
\end{table}

\section{Benchmark Statistics}
\label{app:benchstats}

This section summarizes the evaluation benchmarks used in the main experiments.
The benchmark suite covers static Chroma-backed corpora and live web-backed
retrieval settings, spanning encyclopedic web search, finance filings, patents,
and multi-hop question answering. The split is also used in the main text to
distinguish source-family benchmarks from held-out transfer benchmarks.

Table~\ref{tab:bench-stats} reports the backend, number of test queries, domain,
and task style for each benchmark.

\begin{table}[ht]
\centering
\small
\caption{Benchmark statistics.}
\label{tab:bench-stats}
\resizebox{\textwidth}{!}{
\begin{tabular}{lcccc}
\toprule
\textbf{Benchmark} & \textbf{Backend} & \textbf{Test queries} & \textbf{Domain} & \textbf{Style} \\
\midrule
BrowseComp+~\cite{chen2025browsecompplus}
    & Chroma (BC+ corpus) & 166 & Web encyclopaedic & Multi-constraint \\
Web~\cite{bashir2026context1}
    & Chroma (web 1.17) & 554 & Web encyclopaedic & Multi-hop \\
Patents~\cite{bashir2026context1,uspto_open_data}
    & Chroma (USPTO 1.18) & 718 & Legal/USPTO & Single-doc heavy \\
SEC~\cite{bashir2026context1,sec_edgar_api}
    & Chroma (SEC 1.4) & 502\footnote{\url{}} & Finance filings & Multi-filing \\
LongSealQA~\cite{pham2025sealqa}
    & Chroma (longsealqa) & 254 & Open web & Multi-hop, long-context \\
Seal0QA~\cite{pham2025sealqa}
    & Web (Serper+Jina) & 111 & Open web & Factoid / multi-hop mix \\
FRAMES~\cite{krishna2024frames}
    & Web (Serper+Jina) & 824 & Open web & Multi-hop \\
HotpotQA-subset~\cite{yang2018hotpotqa}
    & Web (Serper+Jina) & 493 & Open web & 2-hop \\
\bottomrule
\end{tabular}
}
\end{table}

\section{Two-Tier Memory and Working-Memory Rendering}
\label{app:wm-rendering}

\sys uses a two-tier memory: an inner \wm summary that fits in
the prompt and an outer \texttt{doc\_store} (full text of every
chunk ever retrieved) accessible via \texttt{review\_docs}. This
section documents the exact rendering protocol so the reported
behaviour is reproducible.

\paragraph{Working-memory text layout.} Each rendered \wm block
contains, in order:
\begin{enumerate}[leftmargin=1.4em,itemsep=2pt,topsep=2pt]
\item Header: \texttt{== Working Memory (summarising turns
0--$n$) ==} followed by the original query.
\item Curated set: rendered grouped by importance level
(\texttt{very\_high} first, \texttt{high}, \texttt{fair},
\texttt{low} last), each entry showing the doc ID and a
$120$-character snippet.
\item Document pool: the most recent $50$ uncurated docs are
shown with snippets; older uncurated docs are summarised as a
comma-separated ID list (truncated at $30$ IDs with a
\texttt{(+N more)} marker if longer).
\item Search history: the last $12$ entries; entries older than
that are summarised as
\texttt{... ($k$ earlier searches)}.
\item Evidence-graph block (only if $\geq 1$ multi-doc bridge
entity exists): up to $8$ frequent entities listed with their
doc IDs, plus a singleton count.
\item Dedup notice (only if any near-duplicates were
suppressed): \texttt{[Dedup] $k$ near-duplicate chunk(s)
auto-suppressed}.
\end{enumerate}

\paragraph{Per-turn observation layout.} Each observation
rendered to the policy contains, in order: (a) the \wm summary
block; (b) up to \texttt{RECENT\_K=5} prior (action, result)
pairs, with reasoning truncated to $300$ characters for all but
the most recent turn; (c) the freshly produced tool result for
the most recent action. Search-style observations within (c) are
passed through sentence-BM25 compression ($K{=}4$ kept sentences)
and content-fingerprint deduplication. Search results are
truncated to \texttt{SEARCH\_DISPLAY\_LIMIT=10} entries per call;
each per-result snippet is capped at
\texttt{DOC\_SNIPPET\_CHARS=120}.

\paragraph{Context-budget truncation (5-pass progressive
degradation).} Whenever the rendered context exceeds the prompt
token budget ($30{,}720$ tokens, leaving $2{,}048$ for
generation within the $32{,}768$ context limit),
\texttt{render\_context\_within\_budget} applies progressive
degradation in five passes, returning the first that fits:
(1)~normal render; (2)~truncate the document-pool section of the
\wm summary while preserving curated-set, search history, and
evidence graph; (3)~truncate analysis text in older turns to
$100$ chars and the \wm summary itself to $2{,}000$ chars;
(4)~drop the oldest recent (action, result) pair and retry,
iterating; (5)~minimal context (system $+$ tool descriptions $+$
query only). Pass 5 always succeeds because the system prompt is
constant-size. This mechanism guarantees that no rollout fails
due to context overflow, and that the curated set and recent
working memory are the last things to be cut.

\section{Per-Turn Programmatic Nudges (Result Summaries)}
\label{app:nudges}

For full transparency: between each turn's tool result and the
next observation, \sys injects a short \emph{result summary} as
a user message. The summary is generated programmatically (no
LLM call) and combines factual status with conditional prescriptive nudges.
The same summaries are seen by the SFT teacher (GPT-5.4) and
the trained policy at RL and inference time, so the prompt
distribution is identical across all three stages.

\paragraph{Always-on factual lines.}
\texttt{[STATUS]}: tool name, number of docs returned, novel
docs, total pool size; or curate/review status.
\texttt{Curated:} current curated set size and a preview of the
first five doc IDs.

\paragraph{Conditional prescriptive lines (gated by
\texttt{HARNESS\_PRESCRIPTIVE=1}, which was on for all reported
runs).}
\texttt{[ACTION REQUIRED]}: triggered after a search turn with
$\geq 1$ turn since last curate; instructs the agent to curate
before searching again.
\texttt{[WARN]}: triggered on multiple consecutive non-curate
turns or a blank curated set while the pool already has candidates.
\texttt{[TIP]}: triggered when truncation is detected
(suggesting \texttt{read\_document}), when only one tool type
has been used by turn $\geq 3$, or when \texttt{grep\_corpus} or
\texttt{read\_document} have not been used by turn $\geq 4$/$6$.
\texttt{[NEXT]}: short suggestion of the immediate next action.

These nudges are \emph{not} part of the trained policy's
intrinsic capability. They are part of the harness. They are
included in the harness mass we report ablating in
Table~\ref{tab:ablation} (the ``all \sys mechanisms disabled''
row removes them).

\section{Harness Algorithms}
\label{app:harness-algorithms}

This appendix gives the formal harness algorithms used by \sys. The
driver in Algorithm~\ref{alg:harness-driver} is intentionally small:
all mechanism-specific behavior is delegated to explicit subroutines.
The same subroutines are used during teacher trajectory generation,
SFT replay, RL rollout, and inference, so the policy is trained on the
same state interface it uses at test time.

\paragraph{State variables.}
For a query $q$, the harness maintains a document pool $P$, a curated
output set $C$ with cap $M{=}30$, an importance map $I:C\to
\{\texttt{very\_high},\texttt{high},\texttt{fair},\texttt{low}\}$, a
full-text store $D$, an evidence graph $G$, a verification cache $V$,
a search history $H$, a near-duplicate index $U$, and a Boolean flag
\textsc{AutoSeeded}. The policy controls semantic choices (queries,
claims, promotions, removals, and termination); the harness controls
deterministic state maintenance.

\begin{algorithm}[t]
\caption{\sys stage driver.}
\label{alg:harness-driver}
\small
\begin{algorithmic}[1]
\REQUIRE query $q$, tools $\mathcal{T}$, stage $s\in\{\mathrm{teacher},\mathrm{sft},\mathrm{rl},\mathrm{infer}\}$, policy $\pi$ or replay trajectory $\tau$
\ENSURE trajectory record for training stages, or curated set $C$ for inference
\STATE Initialize $P,C,I,D,G,V,H,U\leftarrow\emptyset$ and \textsc{AutoSeeded}$\leftarrow$ false
\STATE Build a per-episode rerank instruction from $q$ and the dataset family
\FOR{$t=1,\ldots,T_{\max}$}
  \STATE $x_t\leftarrow\textsc{RenderContext}(q,P,C,I,D,G,V,H,\text{recent turns})$ \hfill Algorithm~\ref{alg:render}
  \IF{$s=\mathrm{sft}$}
    \STATE Read stored action $a_t$ and stored observation $o_t$ from $\tau$
  \ELSE
    \STATE Sample a structured tool action $a_t\sim\pi(\cdot\mid x_t)$
    \STATE Execute $a_t$ with tools $\mathcal{T}$ to obtain raw observation $o_t$
  \ENDIF
  \IF{$a_t$ is a retrieval or memory-inspection action}
    \STATE $(o_t,P,D,G,U,C,I,\textsc{AutoSeeded})\leftarrow\textsc{ProcessObservation}(a_t,o_t)$ \hfill Algorithm~\ref{alg:obs}
  \ELSIF{$a_t=\curate$}
    \STATE $(C,I)\leftarrow\textsc{Curate}(a_t,C,I,P)$ \hfill Algorithm~\ref{alg:curate}
  \ELSIF{$a_t=\verify$}
    \STATE $V\leftarrow V\cup\textsc{VerifyClaim}(a_t,D)$ \hfill Algorithm~\ref{alg:verify-review}
  \ELSIF{$a_t=\texttt{review\_docs}$}
    \STATE $o_t\leftarrow\textsc{ReviewDocs}(a_t,D)$ \hfill Algorithm~\ref{alg:verify-review}
  \ELSIF{$a_t=\texttt{end\_search}$}
    \STATE \textbf{break}
  \ENDIF
  \STATE Update search history $H$, result summary, token marker, and a working-memory snapshot
\ENDFOR
\STATE Apply the stage-specific finalizer in Algorithm~\ref{alg:stages}
\end{algorithmic}
\end{algorithm}

\begin{algorithm}[t]
\caption{Observation processing and working-memory update.}
\label{alg:obs}
\small
\begin{algorithmic}[1]
\REQUIRE action $a$, raw observation $o$, ranked IDs $R$, query text $q$, state $(P,D,G,U,C,I,\textsc{AutoSeeded})$
\ENSURE processed observation and updated memory state
\IF{$a\in\{\texttt{search\_corpus},\texttt{fan\_out\_search},\texttt{grep\_corpus}\}$}
  \STATE Split each returned chunk into sentences and keep the top $K{=}4$ BM25-scored sentences, preserving original order
\ENDIF
\STATE Parse document IDs and document text blocks from $o$
\FOR{each returned chunk $(r,\mathrm{text})$}
  \STATE Normalize $r$ to a document ID $d$ when the dataset uses document-level IDs
  \IF{$d\in P$}
    \STATE skip adding $d$ to the pool, but retain the observation for trajectory accounting
  \ELSIF{$\mathrm{text}$ is a near-duplicate under $U$}
    \STATE suppress $d$ from rendered memory and increment the dedup counter
  \ELSE
    \STATE Add $d$ to $P$ and store full text plus snippet in $D[d]$
    \STATE Extract proper nouns, years, and numeric dates from $\mathrm{text}$ and update $G$
  \ENDIF
\ENDFOR
\IF{$a\in\{\texttt{search\_corpus},\texttt{fan\_out\_search}\}$ and \textsc{AutoSeeded} is false and $R\neq\emptyset$}
  \STATE Add the first $k{=}8$ ranked IDs from $R$ to $C$ with $I(d)=\texttt{fair}$
  \STATE Set \textsc{AutoSeeded}$\leftarrow$ true and append an \texttt{[AUTO-POPULATED]} notice
\ENDIF
\IF{token-budget markers are enabled}
  \STATE Append \texttt{[Context: X/Y]} using the current approximate prompt budget
\ENDIF
\RETURN processed observation and updated state
\end{algorithmic}
\end{algorithm}

\begin{algorithm}[t]
\caption{Importance-aware curation and capacity management.}
\label{alg:curate}
\small
\begin{algorithmic}[1]
\REQUIRE add IDs $A$, remove IDs $R$, proposed importance map $J$, curated set $C$, importance map $I$, pool $P$
\ENSURE updated $(C,I)$ and a capacity-status observation
\STATE Normalize all IDs in $A$, $R$, and $J$ to the dataset's output ID convention
\STATE Remove every $d\in R$ from $C$ and delete $I(d)$
\FOR{each $d\in A$}
  \STATE Let $j\leftarrow J(d)$ if valid, otherwise $j\leftarrow\texttt{fair}$
  \IF{$d\in C$}
    \STATE Re-tag $I(d)\leftarrow j$ and continue
  \ENDIF
  \IF{$|C|<M$}
    \STATE Insert $d$ into $C$ with $I(d)\leftarrow j$
  \ELSE
    \STATE Let $w=\arg\max_{c\in C}\mathrm{rank}(I(c))$, where $\texttt{low}$ has worst rank
    \IF{$\mathrm{rank}(j)<\mathrm{rank}(I(w))$}
      \STATE Evict $w$; insert $d$ with $I(d)\leftarrow j$
    \ELSE
      \STATE Reject $d$ and report a \texttt{[CAPACITY]} marker
    \ENDIF
  \ENDIF
\ENDFOR
\RETURN $(C,I)$ rendered by priority $\texttt{very\_high}\succ\texttt{high}\succ\texttt{fair}\succ\texttt{low}$
\end{algorithmic}
\end{algorithm}

\begin{algorithm}[t]
\caption{Verification and memory review.}
\label{alg:verify-review}
\small
\begin{algorithmic}[1]
\REQUIRE action $a$, full-text store $D$, verification cache $V$
\ENSURE observation $o$ and updated verification cache $V$
\IF{$a=\texttt{review\_docs}(\mathcal{D})$}
  \STATE Return the full text of up to $5$ requested IDs from $D$, without a corpus call
\ELSIF{$a=\verify(\mathcal{D},c)$}
  \FOR{each requested document $d\in\mathcal{D}$}
    \STATE Ask the verifier whether $D[d]$ directly supports every constraint in claim $c$
    \STATE Store $(d,c,\texttt{yes/no},\text{short rationale})$ in $V$
  \ENDFOR
  \STATE Return the per-document yes/no judgments and rationales
\ENDIF
\end{algorithmic}
\end{algorithm}

\begin{algorithm}[t]
\caption{Budget-safe context rendering.}
\label{alg:render}
\small
\begin{algorithmic}[1]
\REQUIRE system prompt, rendered working memory $W$, recent actions and observations, result summaries, token budget $B$
\ENSURE tokenized prompt $x$ with $|x|\leq B$
\STATE Build $W$ from query, curated set grouped by importance, recent uncurated pool, search history, evidence graph, verification cache, dedup notice, and context marker
\STATE Render system message, tool descriptions, query, optional $W$, recent turns, and result summaries
\IF{the tokenized prompt fits within $B$}
  \RETURN prompt
\ENDIF
\STATE Truncate the document-pool section of $W$, preserving curated set, search history, and evidence graph; retry
\STATE If needed, truncate older reasoning to $100$ characters and cap $W$ at $2{,}000$ characters; retry
\STATE If needed, drop the oldest recent action-observation pair one at a time; retry after each drop
\STATE If all retries fail, return the minimal prompt: system message, tool descriptions, and query
\end{algorithmic}
\end{algorithm}

\begin{algorithm}[t]
\caption{Stage-specific finalizers.}
\label{alg:stages}
\small
\begin{algorithmic}[1]
\REQUIRE stage $s$, trajectory $\tau$, pool $P$, curated set $C$, policy $\pi$
\IF{$s=\mathrm{teacher}$}
  \STATE Save all turns, observations, working-memory snapshots, $P$, $C$, document store $D$, and final retrieval metrics
\ELSIF{$s=\mathrm{sft}$}
  \STATE Filter trajectories below the recall threshold
  \STATE Replay each kept trajectory through Algorithms~\ref{alg:obs}--\ref{alg:render}
  \STATE For every turn $t$, train on the supervised pair $\textsc{RenderContext}_t\mapsto a_t$
\ELSIF{$s=\mathrm{rl}$}
  \STATE Score $C$ and $P$ with curated recall, precision, final-answer recall, trajectory recall, and trajectory final-answer recall
  \STATE Compute terminal reward: set quality, trajectory coverage, answer bonus, dense answer shaping, answer-miss penalty, tool-diversity shaping, and turn penalty
  \STATE Update $\pi$ with on-policy CISPO; save checkpoint and rollout summaries
\ELSIF{$s=\mathrm{infer}$}
  \STATE Return the priority-sorted curated set $C$ to the downstream answer generator
\ENDIF
\end{algorithmic}
\end{algorithm}

\section{System Prompts and Templates}
\label{app:prompts}

\subsection{Agent System Prompt (\sys)}
\label{app:syspr-h1}

\begin{promptbox}{Agent system prompt}
\textbf{Role.} You are a retrieval subagent. Find the most relevant
documents from the corpus for the question; do not answer the question
yourself.\\[3pt]
\textbf{Input.}
\texttt{QUERY: [query]}\\[3pt]
\textbf{Tools.} \texttt{fan\_out\_search} runs up to five diverse
queries; \texttt{search\_corpus} runs a single hybrid query;
\texttt{grep\_corpus} performs exact regex matching;
\texttt{read\_document} opens full text; \texttt{review\_docs}
re-renders remembered documents; \curate edits the curated set
(maximum $30$ documents); \verify checks a claim against remembered
documents; \texttt{end\_search} submits the curated set.\\[3pt]
\textbf{Loop.} Decompose the query into constraints. Search, curate
immediately, search again from a different angle, inspect or review
uncertain documents, verify multi-constraint evidence, and end when
the curated set is sufficient. The episode has at most $40$ turns.\\[3pt]
\textbf{State rules.} Use importance tags
\{\texttt{very\_high}, \texttt{high}, \texttt{fair}, \texttt{low}\};
\texttt{very\_high} requires prior verification. The first successful
search seeds tentative \texttt{fair} candidates; promote strong
documents and remove weak ones. Use evidence-graph bridge entities as
follow-up signals. Each observation ends with
\texttt{[Context: X/Y]}; wrap up above $75\%$ and call
\texttt{end\_search} above $90\%$.
\end{promptbox}

\subsection{\texttt{verify} Sub-Prompt}
\label{app:verify-prompt}

\verify is implemented as a small per-document LLM call.
\begin{promptbox}{\texttt{verify} sub-prompt}
\textbf{System.} You are a strict document verifier. Given one
\texttt{CLAIM} and one document's full text, answer only
\texttt{yes} or \texttt{no}, followed by a rationale of at most
$20$ words. Answer \texttt{yes} only if the document directly
supports every part of the claim. Answer \texttt{no} if any constraint
is missing or contradicted. Be conservative.\\[3pt]
\textbf{User template.}\\
\texttt{CLAIM: [claim]}\\
\texttt{DOCUMENT: [document text]}\\
\texttt{Answer: yes/no. rationale}
\end{promptbox}

\subsection{Answer-Generator Prompt (modular RAG)}
\label{app:ansgen-prompt}

\begin{promptbox}{Answer-generator prompt}
\textbf{Task.} You are given a question and retrieved documents. Use
only the documents to reason step by step and answer the question.\\[3pt]
\textbf{Question.} \texttt{[query]}\\[3pt]
\textbf{Retrieved documents.}\\
\texttt{[documents]}\\[3pt]
\textbf{Output rule.} End the response with the final answer on the
last line, prefixed exactly with \texttt{Final Answer:}.
\end{promptbox}
For \textbf{Closed-Book}, the prompt omits the documents and the
instruction is changed to ``Using only your own knowledge, reason
step by step and provide your final answer.'' For
\textbf{Naive RAG}, a single BM25$+$dense hybrid query retrieves
the top-$10$ chunks and those replace the curated set.

\subsection{LLM-as-Judge Prompt}
\label{app:judge-prompt}

We use GPT-5.4 at temperature $0$ as the judge, following the
official BrowseComp+ template; we report the strict
\textsc{correct} rate as ``answer accuracy.''
\begin{promptbox}{LLM-as-judge prompt}
\textbf{Task.} Judge whether \texttt{[response]} answers
\texttt{[question]} correctly according to the precise
\texttt{[correct\_answer]}. Do not solve the problem independently or
comment on irrelevant background.\\[3pt]
\textbf{Inputs.}\\
\texttt{[question]: [query]}\\
\texttt{[response]: [generated\_answer]}\\
\texttt{[correct\_answer]: [gold\_answer]}\\[3pt]
\textbf{Return fields.}
\texttt{extracted\_final\_answer} is the exact final answer, or
\texttt{None} if absent. \texttt{reasoning} explains only meaningful
differences. \texttt{correct} is \texttt{yes} when the extracted
answer matches the gold answer, allowing small numerical tolerances,
and \texttt{no} otherwise. \texttt{confidence} is $0$--$100\%$.
\end{promptbox}

\section{SFT Data Generation Recipe}
\label{app:sft-gen}

SFT trajectories are produced by GPT-5.4
(\texttt{tool\_choice=required},
\texttt{max\_completion\_tokens=4096}) running natively inside
the \sys harness. The teacher sees the same system prompt
(Appendix~\ref{app:syspr-h1}), the same observation layout, and
the same tool schemas as the trained policy will see at RL and
inference time.

\paragraph{Teacher agent loop.}
The generation script runs a single-phase loop: at each turn, the
teacher's full conversation history (system prompt $+$ working-
memory text $+$ the most recent $K{=}3$ turn transcripts $+$
result summaries for older turns) is sent to GPT-5.4 with all
$7{+}1$ tool schemas. The teacher selects a tool via
\texttt{tool\_choice=required} and returns structured JSON
arguments including a mandatory \texttt{reasoning} field. The
harness executes the tool against the live Chroma corpus and
returns the observation.

\paragraph{Guidance protocol.} The generation script injects
turn-level guidance into the prompt: search$\to$curate rhythm
enforcement; \verify nudges on multi-constraint queries (after
$\geq 6$ turns and $\geq 3$ curated docs); backtracking triggers
when the last $3$ search-history entries yielded $\leq 1$ new
doc; tool diversity hints if \texttt{grep\_corpus}/
\texttt{read\_document} are unused after $4$ turns; urgency and
curate-aggressiveness messages near the turn cap. This guidance
ensures teacher trajectories exhibit the full tool vocabulary
and the search$\to$curate rhythm.

\paragraph{Per-dataset quotas and filtering.} Per-dataset raw
quotas: BC$+$/300, SEC/250, Patents/150, Web/150, Web-simple/75,
SEC-simple/75 (total ${\sim}1$K raw). After a $0.10$ recall gate
(trajectories with recall $<0.10$ are discarded), the corpus
contains $899$ trajectories. Per-trajectory expansion into
turn-conditional datums yields ${\sim}26$K training examples.

\paragraph{Auto-populate on first search.}
During trajectory generation, the first successful search
triggers \texttt{auto\_populate\_from\_first\_search}, which adds
the top-$k$ reranked results ($k{=}8$) to the curated set at
\texttt{fair} importance. Idempotent via the per-trajectory
\texttt{wm.auto\_populated} flag. The observation appends an
\texttt{[AUTO-POPULATED]} marker instructing the teacher to
promote good docs and remove irrelevant ones.

\paragraph{Observation processing.} Search observations are
processed identically to RL: BM25 sentence compression
(top-$4$ sentences per chunk), content-hash dedup (MinHash-LSH
with $64$ permutations, 5-gram shingles, threshold $0.85$, with
SHA-prefix fallback over a normalised $4{,}000$-char prefix when
\texttt{datasketch} is unavailable), and the
\texttt{[Context: X/Y]} token marker.

\section{Importance-Aware Curation: Algorithmic Detail}
\label{app:importance}

This section details the eviction policy central to the
subtractive-curation mechanism in \sys (§\ref{sec:method:action}).

\paragraph{Importance levels and ordering.}
Every document carries an importance tag $\in
\{\texttt{very\_high}, \texttt{high}, \texttt{fair},
\texttt{low}\}$ with rank $\texttt{very\_high}{=}0 <
\texttt{high}{=}1 < \texttt{fair}{=}2 < \texttt{low}{=}3$ (lower
rank $=$ higher importance). Auto-populated documents enter at
\texttt{fair}.

\paragraph{ID normalisation.} Both \texttt{add\_ids} and
\texttt{importance} keys are normalised by stripping a trailing
\texttt{\_N} chunk suffix \emph{only if} the resulting ID is not
already in the pool. This lets the policy pass either chunk IDs
(\texttt{42982\_3}) or doc IDs (\texttt{42982}) interchangeably,
while preserving doc IDs that legitimately contain underscores.

\paragraph{Eviction protocol.}
When the curated set is at capacity ($M{=}30$) and the agent
issues a \curate call with new \texttt{add\_ids}:
\begin{enumerate}[leftmargin=1.4em,itemsep=2pt,topsep=2pt]
\item For each incoming doc, compute its
\texttt{incoming\_rank} from the provided importance tag
(default \texttt{fair}).
\item Scan the current curated set for the document with the
\emph{worst} (highest) importance rank.
\item If $\texttt{worst\_rank} > \texttt{incoming\_rank}$, evict
the worst-ranked doc and insert the new doc.
\item If no evictable doc exists, the add is rejected; the
status string includes a \texttt{[CAPACITY]} marker listing up
to $5$ rejected IDs.
\end{enumerate}

\paragraph{Re-tagging in place.}
A \curate call with an \texttt{add\_id} already in the curated
set updates the tag in-place without eviction. This allows the
agent to promote a document after a successful \verify call.

\paragraph{Rendering.}
The curated set is rendered grouped by importance level
(\texttt{very\_high} first, \texttt{low} last) with the tag
visible next to each document ID:
\begin{quote}\small\ttfamily
Curated Set (15/30):\\
\hspace*{1em}very\_high:\\
\hspace*{1em}[*] doc\_42982: Tesla 10-K 2023 mentioning SEC fine...\\
\hspace*{1em}high:\\
\hspace*{1em}[*] doc\_11293: Tesla annual report 2023...\\
\hspace*{1em}fair:\\
\hspace*{1em}[*] doc\_55017: General Motors 10-K 2023...
\end{quote}

\section{Content Deduplication: Algorithmic Detail}
\label{app:dedup}

\sys's content-fingerprint dedup is implemented in
\texttt{ultra\_core.ContentDedupTracker}.

\paragraph{Primary path (MinHash-LSH).} When \texttt{datasketch}
is available, each chunk is fingerprinted by tokenising on the
regex \texttt{[a-z0-9]+}, building $5$-gram shingles, and
inserting them into a MinHash with $64$ permutations. The LSH
index uses Jaccard threshold $0.85$. For each new chunk, the
tracker queries the LSH; any match is treated as a near-duplicate
and dropped from the pool but \emph{retained} in the
\texttt{trajectory\_recall} accounting (so the agent is credited
for having encountered the gold information).

\paragraph{Fallback path (SHA prefix).} If \texttt{datasketch}
is unavailable, the tracker computes a SHA-1 of the first
$4{,}000$ characters of the normalised token-stream and compares
against a hash set. This catches exact duplicates only, not
near-duplicates, but introduces no extra dependency.

\paragraph{Diagnostic.} The number of suppressed duplicates is
exposed in the \wm rendering as a \texttt{[Dedup] $k$
near-duplicate chunk(s) auto-suppressed} line, which lets the
policy reason about whether the corpus has heavy duplication.

\section{Component Ablation: Error Analysis}
\label{app:ablation-error}

This appendix unpacks Table~\ref{tab:ablation} at the trajectory
level. For each ablation we identify queries that full \sys
solves (FA Recall ${\geq}0.5$) but the ablated condition fails
on (FA Recall ${=}0$), then compare action mixes and trajectories
on those \emph{paired} queries. All numbers are over the same
$100$ BrowseComp+ test queries; no retraining is involved.

A consistent pattern emerges: removing a single state mechanism
\emph{changes the policy's behaviour} rather than just losing
information. On failing paired queries the share of
\texttt{search\_corpus} actions rises by $3$--$7$ points, while
\texttt{read\_document} and \verify drop by $2$--$6\times$ ---
a retrieval failure, not a curation failure. The trained
policy reverts to a wide, shallow, search-dominated mode and
never reaches the relevant documents in the first place. The
per-mechanism analyses below each illustrate the pattern with
one representative failing query (truncated for length).

\subsection{Per-mechanism failure modes}

\subsubsection{Importance tags}
\label{app:err:importance-tags}
$\Delta$Recall $-4.1\%$, $\Delta$FA $-7.9\%$ relative
($15$ hard fails).
Binary \curate with FIFO eviction loses the
\texttt{very\_high}/\texttt{high}/\texttt{fair}/\texttt{low}
confidence signal over the curated set. On failing queries,
\texttt{read\_document} drops $7\times$ ($4.7\%{\to}0.7\%$),
\verify drops to $0\%$, and \texttt{search\_corpus} climbs to
$94.1\%$ of all actions.

\textbf{Example.} Q893 asks for the birth name of a 1920s-born
actress married to a director-translator
(gold answer: ``Medea Japaridze''). Full \sys reads
\texttt{doc\_3161}, identifies Medea Japaridze on T4, then
re-anchors via her spouse Revaz Tabukashvili (T28, T37, T39)
and saturates curated recall at $1.0$ in $40$ turns. The
ablation issues $19$ \texttt{search\_corpus} calls without a
single \texttt{read\_document} or \verify, never anchors on
``Medea Japaridze'', and ends with curated recall $0.12$ on
\emph{the same} $40$-turn budget. The agent cannot decide which
candidate to follow up on without the importance gradient.

\subsubsection{Sentence-BM25 compression}
\label{app:err:sentence-compress}
$\Delta$Recall $+0.2\%$, $\Delta$FA $-7.0\%$ relative
($13$ hard fails).
Without per-return sentence compression, top-$k$ chunks come
back raw; returns are longer and noisier. The aggregate Recall
delta is $\sim$$0$, but FA Recall drops sharply because the
agent loses the bridge sentences it normally uses to seed
follow-up reads. \texttt{read\_document} on failing queries
drops from $4.1\%$ to $2.6\%$ and \verify from $2.0\%$ to
$0.4\%$.

\textbf{Example.} Q1012 asks for the title of a $2018$--$2023$ EP
whose songs all end in a question mark and were drawn from the
artist's journal entries. Full \sys recovers it in $26$ turns:
T14 \texttt{grep}s for \texttt{"gave my all"} (a phrase from
the artist interview), reads \texttt{doc\_22401} on T22, and
\verify confirms ``Maki's Tanong'' on T24. The ablation reads
\texttt{doc\_77931} on T2 (a near-miss), then issues
$18$ subsequent searches without ever reading another doc;
without compressed bridge sentences from the search returns it
cannot identify the next phrase to query.

\subsubsection{Auto-seed}
\label{app:err:auto-seed}
$\Delta$Recall $-0.3\%$, $\Delta$FA $-6.4\%$ relative
($12$ hard fails).
With $C_0$ empty, the first \curate has no comparison context,
and the importance-tag space is undefined until $C$ fills up.
Aggregate Recall barely moves because the policy still curates
the right number of documents; FA Recall drops sharply because
the policy retains the wrong documents during the cold-start
window.

\textbf{Example.} Q394 asks for an investment-bank founder who
was a Sagittarius rugby player born in the 1920s (gold answer:
Ron Jarden). Full \sys identifies Ron Jarden on T2 with the
auto-seeded curated set already containing related New
Zealand-rugby chunks for comparison, reads \texttt{doc\_1124} on
T4, \verify-confirms on T6, and ends in $8$ turns with FA
recall $1.0$. The ablation goes through $11$
\texttt{search\_corpus} calls before reading
\texttt{doc\_49990} on T18, but with no warm $C_0$ to anchor
importance against, every candidate looks equally plausible and
the trained \verify call on T20 fails because the supporting
chunks were never curated.

\subsubsection{Evidence graph}
\label{app:err:evidence-graph}
$\Delta$Recall $-2.6\%$, $\Delta$FA $-5.4\%$ relative
($10$ hard fails).
With the graph block hidden, the policy can no longer see which
entities bridge documents in $C$. On failing queries it searches
more ($91.8\%$ of actions vs.~$88.9\%$) and \texttt{read\_document}
drops $3\times$, leaving the curated set without the bridging
entity chunks needed to close the chain.

\textbf{Example.} Q754 asks about a youngest-eligible governor
elected in a newly formed territory with a self-deprecating
slogan and a flamboyant predecessor (gold: Fernando de la R\'ua).
Full \sys leverages the evidence graph to follow the
nickname--territory--predecessor chain across multiple
documents, lands on the correct answer by T34 (FA $1.0$). The
ablation knows nothing about which entities co-occur, so on T39
it tries \texttt{"Fernando de la R\'ua"} as a free-text guess
without the structural anchor, never connecting it to the
``baby bottle'' / ``pacifier'' nickname or the territory-status
clue, and ends at FA $0$.

\subsubsection{\texttt{verify} tool}
\label{app:err:verify}
$\Delta$Recall $-3.1\%$, $\Delta$FA $-3.9\%$ relative
($12$ hard fails).
When \verify returns \texttt{unavailable},
\texttt{read\_document} drops from $5.1\%$ to $0.8\%$ on failing
queries and search rises to $94.6\%$.

\textbf{Example.} Q751 asks for the full name of a University of
Arizona management professor with a $2010$ PhD whose
undergraduate university was founded in the 1800s by a specific
person (gold: William J.\ Becker, UCD). Full \sys reads
\texttt{doc\_15523} on T28, then on T32--T36 traces ``William
Becker'' through the ``University College Dublin'' bridge with
\verify never explicitly invoked but \emph{implicitly} usable.
Without \verify the policy has no closed-loop check and rushes
\textbf{to} commit documents to $C$; on T28--T38 it issues ten
unverified searches and never re-reads its own candidate, so
the curated set ends up with no Becker chunk at all.

\subsubsection{\texttt{review\_docs} tool}
\label{app:err:review-docs}
$\Delta$Recall $+2.4\%$, $\Delta$FA $-3.9\%$ relative
($9$ hard fails).
The aggregate Recall delta is mildly positive because the policy
substitutes additional turn-level reads for the bulk sweep, but
FA Recall still drops because the policy cannot economically
scan the curated set when partial information from many
documents must be combined. \texttt{read\_document} drops
$4\times$ ($4.0\%{\to}1.1\%$) on failing queries and search
rises to $93.6\%$.

\textbf{Example.} Q912 asks for the doctorate university of a
researcher named on a 1995--2005 whale-research paper with $12$
students and $4$ project leaders (gold: University of Hawai\textquotesingle i
at Manoa). Full \sys reads \texttt{doc\_33637} on T4, calls
\verify on T6, then uses \texttt{review\_docs} on T7--T13 to
sweep the curated set for ``Adam Frankel'' co-occurring with
``UH Manoa''; ends in $15$ turns. The ablation reads the same
\texttt{doc\_33637} on T2 but with no bulk re-read it loses
track of the candidate after a handful of new searches and
spends T15--T39 thrashing through name-extraction guesses.

\subsubsection{Content-fingerprint dedup}
\label{app:err:content-dedup}
$\Delta$Recall $+4.6\%$, $\Delta$FA $+1.6\%$ relative
($0$ hard fails).
This is the one ablation that nominally improves the metric.
The cause is mechanical: BC+ qrels occasionally contain
near-duplicate gold documents (e.g., a press release and an
article that quotes it), and \sys's MinHash--LSH dedup
($\theta_{\text{Jaccard}}{=}0.85$) sometimes collapses two such
gold variants into one, costing one gold ID per affected query.
Disabling dedup recovers those IDs but inflates the curated
context with redundant chunks. Dedup is by design a
token-budget mechanism, not a recall mechanism; the
answer-accuracy benchmark in Appendix~\ref{sec:answeracc}
prefers the deduped variant in the majority of paired queries
despite the slightly lower curated recall. We report this row
in the table rather than hide it, because the trade-off is the
design intent.

\subsection{All harness mechanisms disabled}
\label{app:err:all-disabled}
$\Delta$Recall $-12.2\%$, $\Delta$FA $-6.4\%$ relative.
With every \sys
mechanism off simultaneously, the policy has the same action
vocabulary but no compact state to reason over. Recall drops to
$0.513$ and FA recall to $0.624$ --- a larger relative Recall
drop than any single ablation. The agent keeps searching,
sweeping a wide
raw-document space, but cannot rank what it has seen. This is
the cleanest evidence in the paper that the gains attributed to
\sys come from the harness rather than from capability implicit
in the underlying model: on the same queries, the trained
checkpoint searches comparably hard but cannot discriminate
among candidates once the harness is removed.

\section{Evaluation Recipe}
\label{app:eval-recipe}

\paragraph{Metric Definitions}
\label{app:metric-defs}
For each query, let $A_q$ denote the annotated \emph{answer-bearing}
documents: documents that directly contain or support the final answer. Let
$R_q$ denote the full annotated relevant set used for search-quality recall.
On benchmarks with extra evidence annotations, $R_q$ may include additional
supporting documents beyond $A_q$; in BrowseComp+ and Web, for example,
the answer-bearing documents are a subset of the broader relevant set. On
benchmarks without separate answer-document annotations, the evaluator sets
$A_q=R_q$.

Let $C_q$ be the final curated set returned by the search subagent, after
normalizing retrieved chunks to document identifiers. Let $P_q$ be the
trajectory pool: all documents encountered anywhere in the episode, again
normalized to document identifiers. We report three recall-oriented metrics:
\[
\mathrm{Recall}(q) = \frac{|C_q \cap R_q|}{|R_q|}, \qquad
\mathrm{FinalAnswerRecall}(q) = \frac{|C_q \cap A_q|}{|A_q|},
\]
\[
\mathrm{TrajectoryRecall}(q) = \frac{|P_q \cap R_q|}{|R_q|}.
\]
Dataset-level scores are macro-averages over queries. For fact-level datasets,
the same definitions apply with facts in place of documents: a fact is counted
as covered if any annotated chunk for that fact is retrieved.

Final-Answer Recall is not mathematically constrained to be above or below
Recall. When $A_q \subset R_q$, the denominators differ: Recall averages over
all relevant annotations, while Final-Answer Recall averages only over
answer-bearing annotations. Thus a system that finds answer documents but misses
many auxiliary supporting documents can have Final-Answer Recall $>$ Recall,
whereas a system that finds supporting documents but misses answer documents can
have Final-Answer Recall $<$ Recall. Trajectory Recall is an upper-bound-style
diagnostic for discovery before final curation; it need not equal final Recall
because the agent may encounter evidence and later omit it from $C_q$.



\paragraph{Reranking for released open-agent harnesses.}
Search-R1 and Tongyi DeepResearch are evaluated with their released agent harnesses, since the harness is part of each method. These harnesses do not natively return an importance-tagged curated set capped at $30$ documents. We therefore standardize their final outputs with a shared reranker. For each query, we collect all unique documents encountered during the episode and normalize chunks to document identifiers. If the trajectory pool contains at most $30$ documents, we use the full pool as the curated set $C_q$ without reranking. Otherwise, we rerank the trajectory pool with the shared reranker and keep the top-$30$ documents as $C_q$. Recall and Final-Answer Recall are computed from $C_q$, while Trajectory Recall is computed from the full trajectory pool $P_q$ before reranking.

For Search-R1, this final-set construction usually does not change the
retrieval outcome much. Although Search-R1 is evaluated under the same protocol,
its released harness typically performs only four to five text-search rounds in
our runs. This is consistent with its training setup, which optimizes short
search-and-answer trajectories rather than long-horizon evidence collection and
curation. As a result, Search-R1 rarely accumulates a trajectory pool much
larger than the final document budget. Table~\ref{tab:search_quality} reflects
this behavior: Search-R1's Recall and Trajectory Recall are nearly identical
across benchmarks, suggesting that there is little additional evidence in the
trajectory pool to recover through post-hoc curation.

\paragraph{Answer-accuracy pipeline (modular RAG protocol).} To
decouple retrieval from generation, each subagent's curated set is
forwarded to a frozen frontier generator that sees \emph{only} the
query and curated documents, with no trajectory, intermediate
reasoning, or tool-use history. For each
(subagent~$\times$~benchmark) condition, we read the saved trajectory
JSON, extract the curated document identifiers and corresponding chunk
text from observation logs, format with the prompt in
Appendix~\ref{app:ansgen-prompt}, and dispatch to four generators:
GPT-5.4, Sonnet-4.6, Opus-4.6, and Kimi-K2.5. We use the same
benchmark instances as the main evaluation. Closed-Book and
Naive-RAG (top-$10$ BM25$+$dense) anchor the protocol. The judge
(Appendix~\ref{app:judge-prompt}) is GPT-5.4 at $T{=}0$ with the
BrowseComp+ template; we report the strict \textsc{correct} rate.
Every (query, generator, generated-answer, verdict) tuple is saved.

\section{Modular RAG: Curated Sets Yield Higher Answer Accuracy}
\label{sec:answeracc}

Search quality measures what the subagent returns; what ultimately
matters is the answer the downstream system gives. Under the modular
RAG protocol described above, the answer-generator LLM sees only the
curated set, never the trajectory. Thus, any accuracy difference
between subagent rows is attributable to the curated set alone.

\begin{figure}[h]
\centering
\includegraphics[width=\textwidth]{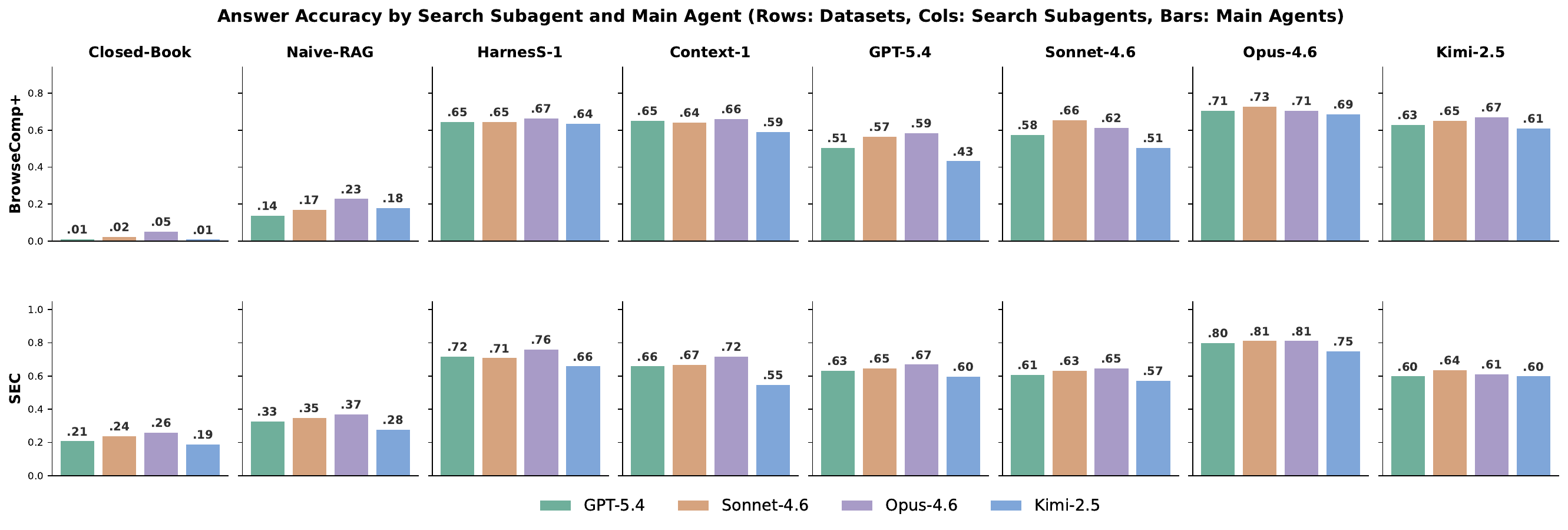}
\caption{\textbf{Answer accuracy} (\%) under the modular RAG
protocol. Each subplot is a benchmark; each group of bars is a
search method (or baseline); each bar colour is a different
frontier answer-generator. \emph{Closed-Book}: generator answers
from parametric knowledge alone. \emph{Naive RAG}: single
BM25$+$dense top-$10$ chunk retrieval, no agentic search. All
other groups: an agentic search subagent curates the document
set; the generator receives \emph{only} query and curated
documents, no trajectory. Judging follows the official
BrowseComp+ template (GPT-5.4, $T{=}0$).}
\label{fig:answeracc}
\end{figure}

Three patterns are visible in Figure~\ref{fig:answeracc}. First,
Closed-Book and Naive RAG are weak on the hardest displayed benchmark
(BC+), confirming that these tasks require deep agentic search;
parametric knowledge and single-hop retrieval are insufficient. Second,
the row ordering across subagents broadly tracks the search-quality
ordering from Table~\ref{tab:search_quality}: better curated sets yield
higher answer accuracy across all four generators. Third, column
variation within a row is modest for strong searchers, since every
frontier generator can converge on the right answer when given the
right documents, and larger for weaker searchers, where the generator
must compensate for missing evidence. Because the generator never sees
the trajectory, the rank ordering of subagent rows is a clean test of
curated-set quality: \sys's curated sets dominate the open subagents in
the reported modular RAG settings.

\section{Harness as Confound: Same LLM, Different Harness}
\label{app:harness-confound}

A natural concern is that all frontier baselines are evaluated
under the Context-1 harness, not under \sys's harness. This is
deliberate: we compare \emph{trained agents} (Context-1 RL,
\sys RL) each running in their own harness, while frontier LLMs
serve as untrained baselines using Context-1's harness as a
common reference environment.

To quantify the harness contribution alone, we ran GPT-5.4 under
three harness conditions on three in-domain benchmarks:
\textbf{Naive Search-Add} (minimal loop, no curation, no
compression, no dedup); \textbf{Context-1 Harness} (production
Context-1 environment); \textbf{\sys Harness} (full \sys
environment).

\begin{figure}[h]
\centering
\includegraphics[width=0.7\textwidth]{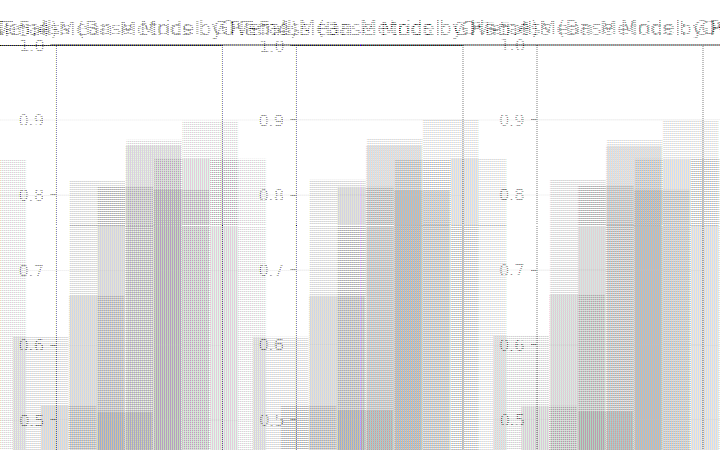}
\caption{\textbf{Same LLM (GPT-5.4), different harness.} All
four retrieval metrics improve monotonically as the harness
grows richer: naive search-add $\to$ Context-1 harness $\to$
\sys harness. The harness contributes a $+4$ point recall gain
on top of Context-1.}
\label{fig:harness-confound}
\end{figure}

The progression is monotonically improving. Curated Recall jumps
from $0.511$ (naive) to $0.807$ (Context-1) to $0.849$ (\sys);
Final-Answer Recall from $0.612$ to $0.821$ to $0.876$. A $+4.2$
point recall gain is available to GPT-5.4 simply by switching
from Context-1's harness to \sys's, \emph{without any RL
training}. This validates the central thesis: the harness is a
compute-allocation mechanism that materially changes how much a fixed
model can discover.

\newpage
\section{Qualitative Case Studies}
\label{app:case-studies}

The following abridged trajectories show where the harness changes the
search process: auto-seeding creates an editable state, curation
promotes evidence, exact search recovers needles, backtracking removes
false leads, and full-document reading resolves the final answer.
Routine prompt boilerplate, redundant curation calls, false-positive
documents, and repeated metadata are omitted; the unabridged logs are
retained separately.

\subsection{Case Study A: Stateful Search over a Drug-Approval Query}
\label{app:case-bc643}

\begin{promptbox}{BrowseComp+ query}
\textbf{Query.} An agent is not indicated for patients who weigh 30 kg or are 10 years old. It can be taken via oral administration, and one of the side effects reported is nausea. A study from 2022 mentions that its half-life is less than 10 hours and that the liver primarily eliminates it. The study is authored by two people, published in December, and features 31 references. The application for tablets containing the agent was received on March 23, and it was approved eight months later by a letter. What is the full name (as mentioned in the letter) of the Regulatory Project Manager to be called in case of any questions?\\[2pt]
\textbf{Gold answer.} Alicia Moruf\\[2pt]
\textbf{Outcome.} 18 turns, tool diversity $6$, curated recall $1.0$, final-answer recall $1.0$.
\end{promptbox}

\noindent\textbf{Why this case is interesting.}
The query is a regulatory needle: the agent must connect label constraints, a pharmacology study, an NDA package, and an approval letter naming the project manager. The trajectory shows the harness doing useful bookkeeping rather than merely storing a transcript. Auto-seeding creates the initial regulatory evidence set, curation separates maribavir and FDA-package evidence from distractors, exact search recovers from brittle phrase matching, and full-document reading exposes the answer sentence.

\noindent\textbf{How the trajectory is abridged.}
The full log keeps every token. Here we omit the system prompt, repeated curation confirmations, low-value false-positive documents, and two no-op verify/review calls caused by chunk-id mismatch. The retained turns preserve the model reasoning, exact tool calls, decisive returns, and state snapshots needed to see the harness at work.

\paragraph{Trajectory overview.}
{\small
\setlength{\tabcolsep}{3pt}
\renewcommand{\arraystretch}{1.08}
\begin{longtable}{p{0.055\textwidth}p{0.12\textwidth}p{0.36\textwidth}p{0.36\textwidth}}
\toprule
Turn & Tool & Policy move & Harness state / why it matters \\
\midrule
\endfirsthead
\toprule
Turn & Tool & Policy move & Harness state / why it matters \\
\midrule
\endhead
0 & \CaseTool{toolSearch}{search} & Search all medical clues at once & auto-seed creates the first editable curated set (6 docs, because only 6 were returned) \\
1 & \CaseTool{toolCurate}{curate} & Promote Livtencity and pharmacology candidates & curation separates strong maribavir evidence from low-value distractors \\
2 & \CaseTool{toolSearch}{search} & Search the 2022 maribavir study & connects half-life, hepatic metabolism, two authors, and references \\
3 & \CaseTool{toolCurate}{curate} & Promote study and label chunks & working memory now contains drug identity plus PK evidence \\
4 & \CaseTool{toolSearch}{search} & Search FDA approval-letter wording & same document family reappears with NDA context \\
5 & \CaseTool{toolCurate}{curate} & Consolidate FDA package and study chunks & curated state keeps both regulatory and pharmacology evidence \\
6 & \CaseTool{toolSearch}{search} & Search by regulatory-manager wording & narrows to approval package chunks \\
7 & \CaseTool{toolCurate}{curate} & Add later FDA-letter chunks & candidate pool now includes the likely answer document \\
8 & \CaseTool{toolGrep}{grep} & Exact grep for phrase & a failed exact match prevents over-trusting a brittle pattern \\
10 & \CaseTool{toolSearch}{search} & Search by NDA number & needle appears: Alicia Moruf in the approval letter \\
11 & \CaseTool{toolCurate}{curate} & Promote NDA-letter chunk & final evidence enters curated memory \\
12 & \CaseTool{toolRead}{read} & Read full approval-letter chunk & outer-tier memory exposes the decisive text \\
13 & \CaseTool{toolCurate}{curate} & Reaffirm final evidence & curated set keeps approval letter high \\
\bottomrule
\end{longtable}
}
\subsubsection*{\textcolor{toolSearch!80!black}{Phase 1 - Auto-seeded working memory from a broad medical search}}
\begin{trajectorylisting}{toolNeutral}{Reasoning excerpt}
Key constraints: agent not indicated for patients >=30kg or 10y; oral administration; side effect nausea; 2022 study mentions half-life <10h, primarily liver elimination; study by two authors, December, 31 references; tablet application received 23 Mar; approved 8 months later via letter identifying a Regulatory Project Manager. Likely a US FDA approval letter for an oral drug for weight >=30kg/age >=10y contraindication. The unique medicine clue is the age/weight contraindication plus nausea side effect. Search angle: distinct clinical label and side effect plus half-life/liver elimination trait.
\end{trajectorylisting}

\begin{trajectorylisting}{toolSearch}{Exact tool call}
[
  {
    "tool": "search_corpus",
    "params": {
      "query": "oral contraindicated for under 30 kg age 10 nausea half-life less than 10 hours liver eliminated 2022 study December 31 references FDA approval letter Regulatory Project Manager March 23 seventeen Feb 2023 app tablets received March 23 approved eight months letter"
    }
  }
]
\end{trajectorylisting}

\begin{trajectorylisting}{toolSearch}{Search return: first hit creates the regulatory anchor}
# DOCUMENT ID: 59631_0 (743 tokens) CENTER FOR DRUG EVALUATION AND RESEARCH Approval Package for: APPLICATION NUMBER: 215596Orig1s000 Trade Name: LIVTENCITY oral tablets Generic or Proper Name: maribavir Sponsor: Takeda Pharmaceuticals USA, Inc. Approval Date: November 23, 2021 Indication: For the treatment of adults and pediatric patients 12 years of age and older and weighing at least 35 kg with post-transplant cytomegalovirus (CMV) infection/disease that is refractory to treatment (with or without genotypic resistance) with ganciclovir, valganciclovir, cidofovir or foscarnet. CENTER FOR DRUG EVALUATION AND RESEARCH 215596Orig1s000 CONTENTS Reviews / Information Included in this NDA Review. Approval Letter X Other Action Letters Labeling X REMS Officer/Employee List X Integrated Review(s) X Product Quality Review(s) X Other Reviews X Risk Assessment and Risk Mitigation Review(s) X Proprietary Name Review(s) X Administrative/Correspondence Document(s) X CENTER FOR DRUG EVALUATION AND RESEARCH APPLICATION NUMBER: 215596Orig1s000 APPROVAL LETTER NDA 215596 NDA APPROVAL Takeda Pharmaceuticals USA, Inc. Attention: Michael A. Cronin, PharmD Director, Global Regulatory Affairs 650 ...
\end{trajectorylisting}

\begin{trajectorylisting}{toolCurate}{Curation update}
Curated set updated (6/30): 25814[high], 59631[high], 15546[fair], 50858[fair], 66387[low], 97801[low]
\end{trajectorylisting}

\begin{trajectorylisting}{toolNeutral}{State after turn 1}
pool_size: 6
curated_size: 6/30
importance: very_high=0, high=2, fair=2, low=2
top_curated:
  - 25814 [high]
  - 59631 [high]
  - 15546 [fair]
  - 50858 [fair]
  - 66387 [low]
  - 97801 [low]
\end{trajectorylisting}

\subsubsection*{\textcolor{toolSearch!80!black}{Phase 2 - The policy pivots to pharmacokinetic evidence}}
\begin{trajectorylisting}{toolNeutral}{Reasoning excerpt}
The first search hit a known FDA submission doc for Livtencity (maribavir). That fits oral administration, nausea side effect, and contraindication at 30kg/10y, but the drug class/youth contraindication constraint is strong. We need the 2022 study page and the approval letter identifying the Regulatory Project Manager. The search angle using the study constraints is still relevant; I'll first search for the 2022 study on maribavir to bridge to the approval letter.
\end{trajectorylisting}

\begin{trajectorylisting}{toolSearch}{Exact tool call}
[
  {
    "tool": "search_corpus",
    "params": {
      "query": "maribavir 2022 study liver eliminated half-life less than 10 hours oral migration 31 references December two authors"
    }
  }
]
\end{trajectorylisting}

\begin{trajectorylisting}{toolSearch}{Search return: study and label evidence}
# DOCUMENT ID: 83797_4 (723 tokens) As a postmarketing commitment, the FDA is requiring the manufacturer to conduct phenotypic analysis of maribavir against human CMV. Substitutions designated as a high priority for analysis are pUL97 M460I/T, A594E/P/T/V, L595F/W, and C603R/W/Y. Substitutions designated as medium priority are pUL97 A440V, V466M, A591V, E596G, K599E, and C607F/Y. Substitutions designated as low priority are pUL97 P132L, L405P, C518Y, I610T, and A613V; pUL27 P10L, N289D, H29Y, D298G, N300G, P307L, V310A, D351N, and I367V; and pUL54 S290R and K475S.18 Maribavir does not prolong the QT interval to a clinically important extent.1,19 Pharmacokinetics Following oral administration, median T max is 1 to 3 hours.1,19,20 Plasma exposure (C max and AUC) increased approximately dose proportionally following a single dose of 50 to 1600 mg and multiple doses of up to 2400 mg/day.1,20 Compared with administration in the fasted state, administration with a moderate-fat meal resulted in a small reduction in AUC 0-infinity (geometric mean ratio, 0.864; 90% CI, 0.804-0.929) and C max (0.722; 90% CI, 0.656-0.793). Steady state is reached within 2 days with twice-daily dosing.1 Mean steady-state volume of distribution is 27.3 L. Maribavir is 98% plasma protein bound. The blood-to-plasma ratio is 1.37.1,20 Maribavir does not appear to cross the blood-brain barrier.9 Maribavir, the parent drug, is responsible for the ...

# DOCUMENT ID: 17183_8 (746 tokens) due to the parent drug. Following oral administration, plasma maribavir exposure (Cmax and AUC) increased approximately dose-proportionally following a single dose of 50 to 1600 mg (0.125 to four times the recommended dose) and multiple doses up to 2400 mg per day (three times the recommended daily dose). Maribavir PK is time-independent. With twice-daily dosing, steady state is reached within 2 days, with mean accumulation ratios of Cmax and AUC ranging from 1.37 to 1.47. The pharmacokinetic properties of maribavir following administration of LIVTENCITY are displayed in Table 4. The multiple-dose pharmacokinetic parameters are provided in Table 5. 9 Reference ID: 4893709 This label may not be the latest approved by FDA. For current labeling information, please visit Table 4: Pharmacokinetic Properties of Maribavir Absorptiona Tmax (h), median 1.0 to 3.0 Distribution Mean apparent steady-state volume of distribution (Vss, L) 27.3 % bound to human plasma proteins 98.0 across the concentration range of 0.05-200 ug/mL Blood-to plasma ratio 1.37 Elimination Major route of elimination Hepatic metabolism Half-life (t1/2) in transplant patients (h), mean 4.32 Oral clearance (CL/F) in transplant patients (L/h), mean 2.85 Metabolism Metabolic pathwaysb CYP3A4 (major) and CYP1A2 (minor) Excretion % of dose excreted as total 14C (unchanged drug) in urinec 61 (<2) % of dose excreted as total 14C ...

# DOCUMENT ID: 73877_2 (477 tokens) B: May be acceptable. Either animal studies show no risk but human studies not available or animal studies showed minor risks and human studies done and showed no risk. C: Use with caution if benefits outweigh risks. Animal studies show risk and human studies not available or neither animal nor human studies done. D: Use in LIFE-THREATENING emergencies when no safer drug available. Positive evidence of human fetal risk. X: Do not use in pregnancy. Risks involved outweigh potential benefits. Safer alternatives exist. NA: Information not available.Pharmacology Mechanism of Action Antiviral activity mediated by competitive inhibition of the protein kinase activity of human CMV enzyme pUL97, which inhibits protein phosphorylation Maribavir inhibited human CMV replication in virus yield reduction, DNA hybridization, and plaque reduction assays in human lung fibroblast cell line, human embryonic kidney, and human foreskin fibroblast cells Absorption Peak plasma time: 1-3 hr AUC: 128 ug*hr/mL Peak plasma concentration: 17.2 ug*hr/mL Distribution Vd (steady-state): 27.3 L Plasma bound: 98% Blood-to-plasma ratio: 1.37 Metabolism Metabolized by CYP3A4 (major) and CYP1A2 minor Elimination Clearance: 2.85 L/hr Half-life: 4.32 hr Excretion: Urine (61% [<2% unchanged]); feces (14% [5.7%]) Administration Oral Administration Take with or without food Storage Store at 20-25oC (68-77oF); excursions ...
\end{trajectorylisting}

\begin{trajectorylisting}{toolCurate}{Curation update}
Curated set updated (9/30): 17183[high], 25814[high], 59631[high], 83797[high], 15546[fair], 50858[fair], 73877[fair], 66387[low], 97801[low]
\end{trajectorylisting}

\begin{trajectorylisting}{toolNeutral}{State after turn 3}
pool_size: 9
curated_size: 9/30
importance: very_high=0, high=4, fair=3, low=2
top_curated:
  - 17183 [high]
  - 25814 [high]
  - 59631 [high]
  - 83797 [high]
  - 15546 [fair]
  - 50858 [fair]
  - 73877 [fair]
  - 66387 [low]
  - 97801 [low]
\end{trajectorylisting}

\subsubsection*{\textcolor{toolGrep!80!black}{Phase 3 - Exact search misses, so state prevents a blind loop}}
\begin{trajectorylisting}{toolGrep}{Exact grep call}
[
  {
    "tool": "grep_corpus",
    "params": {
      "pattern": "Maribavir.*Regulatory Project Manager"
    }
  }
]
\end{trajectorylisting}

\begin{trajectorylisting}{toolGrep}{Grep return}
No results found
[Context: 11553/30720]
\end{trajectorylisting}

\begin{trajectorylisting}{toolSearch}{Recovery search by NDA number}
[
  {
    "tool": "search_corpus",
    "params": {
      "query": "NDA 215596 letter to maribavir regulatory project manager NDA 215596"
    }
  }
]
\end{trajectorylisting}

\begin{trajectorylisting}{toolSearch}{Decisive return around answer}
[... clipped before ...]
EPORTING REQUIREMENTS We remind you that you must comply with reporting requirements for an approved NDA (21 CFR 314.80 and 314.81). POST APPROVAL FEEDBACK MEETING New molecular entities qualify for a post approval feedback meeting. Such meetings are used to discuss the quality of the application and to evaluate the communication process during drug development and marketing application review. The purpose is to 4 For the most recent version of a guidance, check the FDA guidance web page at 5 6 Reference ID: 4893709 NDA 215596 Page 6 U.S. Food and Drug Administration Silver Spring, MD 20993  learn from successful aspects of the review process and to identify areas that could benefit from improvement. If you would like to have such a meeting with us, call the Regulatory Project Manager for this application. If you have any questions, call Alicia Moruf, PharmD, MPH, RAC-US, Regulatory Project Manager, at 301-796-3953. Sincerely, {See appended electronic signature page} John Farley, MD, MPH Director Office of Infectious Diseases Office of New Drugs Center for Drug Evaluation and Research ENCLOSURE(S): - Content of Labeling o Prescribing Information o Patient Package Insert - Carton and Container Labeling Reference ID: 4893709 --------------------------------------------------------------------------------------------This is a representation of an electronic record that was signed electronically. Following this are manifestations of any and all electronic signatures for this electronic record.

# DOCUMENT ID: 73877_0 (719 tokens) 
---
title: maribavir (Rx)
date: 2025-01-01
---
Dosing & Uses

Dosage Forms & Strengths

tablet

- 200mg

Cytomegalovirus Infection

Indicated for t
[... clipped after ...]
\end{trajectorylisting}

\begin{trajectorylisting}{toolCurate}{Curation update}
Curated set updated (10/30): 17183[high], 25814[high], 15546[fair], 50858[fair], 59631[fair], 73877[fair], 66387[low], 72434[low], 83797[low], 97801[low]
\end{trajectorylisting}

\begin{trajectorylisting}{toolNeutral}{State after turn 11}
pool_size: 11
curated_size: 10/30
importance: very_high=0, high=2, fair=4, low=4
top_curated:
  - 17183 [high]
  - 25814 [high]
  - 15546 [fair]
  - 50858 [fair]
  - 59631 [fair]
  - 73877 [fair]
  - 66387 [low]
  - 72434 [low]
  - 83797 [low]
  - 97801 [low]
\end{trajectorylisting}

\subsubsection*{\textcolor{toolRead!80!black}{Phase 4 - Full-document reading confirms the final evidence}}
\begin{trajectorylisting}{toolRead}{Exact read call}
[
  {
    "tool": "read_document",
    "params": {
      "doc_id": "59631_3"
    }
  }
]
\end{trajectorylisting}

\begin{trajectorylisting}{toolRead}{Full-document excerpt around answer}
[... clipped before ...]
ilable at FDA.gov.5 Information and Instructions for completing the form can be found at FDA.gov.6 REPORTING REQUIREMENTS We remind you that you must comply with reporting requirements for an approved NDA (21 CFR 314.80 and 314.81). POST APPROVAL FEEDBACK MEETING New molecular entities qualify for a post approval feedback meeting. Such meetings are used to discuss the quality of the application and to evaluate the communication process during drug development and marketing application review. The purpose is to 4 For the most recent version of a guidance, check the FDA guidance web page at 5 6 Reference ID: 4893709 NDA 215596 Page 6 U.S. Food and Drug Administration Silver Spring, MD 20993  learn from successful aspects of the review process and to identify areas that could benefit from improvement. If you would like to have such a meeting with us, call the Regulatory Project Manager for this application. If you have any questions, call Alicia Moruf, PharmD, MPH, RAC-US, Regulatory Project Manager, at 301-796-3953. Sincerely, {See appended electronic signature page} John Farley, MD, MPH Director Office of Infectious Diseases Office of New Drugs Center for Drug Evaluation and Research ENCLOSURE(S): - Content of Labeling o Prescribing Information o Patient Package Insert - Carton and Container Labeling Reference ID: 4893709 --------------------------------------------------------------------------------------------This is a representation of an electronic record that was signed electronically. Following this are manifestations of any and all electronic signatures for this electronic record.--------------------------------------------------------------------------------------------/s/ ------------------------------------------------------------JOHN J FARLEY 11/23/2021 03:34:13 PM Signature Page 1 of 1 Reference ID: 4893709 (
[Context: 10329/30720]
\end{trajectorylisting}

\begin{trajectorylisting}{toolCurate}{Final curation update}
Curated set updated (10/30): 17183[high], 25814[high], 59631[high], 15546[fair], 50858[fair], 73877[fair], 66387[low], 72434[low], 83797[low], 97801[low]
\end{trajectorylisting}

\begin{trajectorylisting}{toolNeutral}{State after turn 13}
pool_size: 11
curated_size: 10/30
importance: very_high=0, high=3, fair=3, low=4
top_curated:
  - 17183 [high]
  - 25814 [high]
  - 59631 [high]
  - 15546 [fair]
  - 50858 [fair]
  - 73877 [fair]
  - 66387 [low]
  - 72434 [low]
  - 83797 [low]
  - 97801 [low]
\end{trajectorylisting}

\clearpage
\subsection{Case Study C: Exact-Date Recovery in SEC Filings}
\label{app:case-sec497}

\begin{promptbox}{SEC query}
\textbf{Query.} There is a company where a senior financial executive with over two decades of consumer packaged goods experience joined as a senior vice president in the early spring of the mid-2020s. Around the early summer of the same year, a different long-serving financial executive concluded their tenure after announcing their decision to step down several months earlier. That same date marked when the spring hire assumed the top financial role, succeeding the retiring executive.
On what date did this leadership transition in the financial executive role become effective?\\[2pt]
\textbf{Gold answer.} July 3, 2025\\[2pt]
\textbf{Outcome.} 9 turns, tool diversity $5$, curated recall $1.0$, final-answer recall $1.0$.
\end{promptbox}

\noindent\textbf{Why this case matters.}
SEC queries often fail when the agent treats filings as isolated snippets. This trajectory shows the source-family behavior behind the SEC gains: auto-seeding starts inside the right filing family, curation preserves the relevant 8-Ks, \texttt{read\_document} exposes the full executive-transition language, and \texttt{grep\_corpus} verifies the exact effective date across later filings.

\paragraph{Trajectory overview.}
{\small
\setlength{\tabcolsep}{3pt}
\renewcommand{\arraystretch}{1.08}
\begin{longtable}{p{0.055\textwidth}p{0.12\textwidth}p{0.36\textwidth}p{0.36\textwidth}}
\toprule
Turn & Tool & Policy move & Harness state / why it matters \\
\midrule
\endfirsthead
\toprule
Turn & Tool & Policy move & Harness state / why it matters \\
\midrule
\endhead
0 & \CaseTool{toolSearch}{search} & Search executive-transition clues in SEC filings & first hit already binds company, outgoing CFO, incoming CFO, and effective date \\
1 & \CaseTool{toolCurate}{curate} & Promote the relevant 8-Ks & curated set separates the correct filing family from other finance hits \\
3 & \CaseTool{toolRead}{read} & Read the May 8-K & full text resolves succession details rather than relying on snippets \\
4 & \CaseTool{toolCurate}{curate} & Keep the read filing high & state retains the source document for final answer recall \\
6 & \CaseTool{toolGrep}{grep} & Grep the exact effective date & regex recovers later proxy mentions of the same transition \\
7 & \CaseTool{toolCurate}{curate} & Promote corroborating proxy chunks & final state keeps multiple filings about the CFO transition \\
\bottomrule
\end{longtable}
}
\subsubsection*{\textcolor{toolSearch!80!black}{Phase 1 - Auto-seeding finds the right company and date}}
\begin{trajectorylisting}{toolSearch}{Exact search call}
[
  {
    "tool": "search_corpus",
    "params": {
      "query": "spring 2024 senior vice president food consumer packaged goods 20 years CPG effective CFO transition summer 2024 retire step down senior financial executive announced decision months earlier"
    }
  }
]
\end{trajectorylisting}

\begin{trajectorylisting}{toolSearch}{Decisive first return}
# DOCUMENT ID: 0001104659-25-006698_0 (387 tokens) 
Company: Simply Good Foods Co
Filing Date: 2025-01-28
Form: 8-K
Item: Item 5.02
Chunk 0
---
Item 5.02 Departure of Directors or Certain Officers; Election of
Directors; Appointment of Certain Officers; Compensatory Arrangements of Certain Officers.

(b) On January 27, 2025, Shaun P. Mara, Chief Financial Officer
of The Simply Good Foods Company (the " Company"), notified the Company of his decision to retire effective July 3, 2025.
Mr. Mara will remain as Chief Financial Officer through the date of his retirement, at which time he will resign from all positions
with the Company and its subsidiaries.

In connection with Mr. Mara's retirement, on January 28,
2025, the Company announced the hiring of Mr. Christopher J. Bealer as Senior Vice President of Finance of the Company, as of
April 1, 2025. Mr. Bealer is expected to succeed Mr. Mara as Chief Financial Officer upon Mr. Mara's retirement
on July 3, 2025.

Mr. Bealer has almost 23 years of experience in consumer packaged
goods and consumer durables in North America and global markets. In his most recent executive leadership position as EVP Corporate Controller
at Reckitt Benckiser Group PLC (" Reckitt"), Mr. Bealer leads a global team of over 600 employees accountable for global
controls, financial shared services, group reporting and accounting, as well as leading the functional operating model transformation.
Prior to his current and prior roles at Reckitt, Mr. Bealer served as Regional CFO for Whirlpool Corporation for both the North American
and Europe, Middle East and Africa (EME
[... clipped after ...]
\end{trajectorylisting}

\begin{trajectorylisting}{toolCurate}{Curation update}
Curated set updated (7/30): 0001104659-25-006698_0[high], 0001104659-25-053287_0[high], 0001104659-25-121934_20[high], 0000909832-25-000101_19[fair], 0001364954-25-000039_57[fair], 0001673985-25-000053_24[fair], 0001530950-25-000260_18[low]
\end{trajectorylisting}

\begin{trajectorylisting}{toolNeutral}{State after turn 1}
pool_size: 7
curated_size: 7/30
importance: very_high=0, high=3, fair=3, low=1
top_curated:
  - 0001104659-25-006698_0 [high]
  - 0001104659-25-053287_0 [high]
  - 0001104659-25-121934_20 [high]
  - 0000909832-25-000101_19 [fair]
  - 0001364954-25-000039_57 [fair]
  - 0001673985-25-000053_24 [fair]
  - 0001530950-25-000260_18 [low]
\end{trajectorylisting}

\subsubsection*{\textcolor{toolRead!80!black}{Phase 2 - Full-document reading keeps the answer grounded}}
\begin{trajectorylisting}{toolRead}{Exact read call}
[
  {
    "tool": "read_document",
    "params": {
      "doc_id": "0001104659-25-053287_0"
    }
  }
]
\end{trajectorylisting}

\begin{trajectorylisting}{toolRead}{Full-document excerpt around transition}
# Document (1500 tokens)
Company: Simply Good Foods Co
Filing Date: 2025-05-28
Form: 8-K
Item: Item 5.02
Chunk 0
---
Item 5.02

Departure of Directors or Certain Officers; Election of Directors; Appointment of Certain Officers; Compensatory Arrangements of Certain Officers.

Appointment of New CFO

In connection with the previously announced Chief Financial Officer
transition, on May 23, 2025, The Simply Good Foods Company (the " Company") appointed Christopher J. Bealer as the Company's
Chief Financial Officer effective July 3, 2025. Mr. Bealer is currently serving as the Company's Senior Vice President,
Finance having joined the Company on April 1, 2025. As Chief Financial Officer, Mr. Bealer will continue to report to Geoff
E. Tanner, the Company's President and Chief Executive Officer.

Mr. Bealer, age 48, has almost 23 years of experience in consumer-packaged
goods and consumer durables in North America and Global markets. Mr. Bealer was most recently EVP Corporate Controller at Reckitt
Benckiser Group PLC (" Reckitt") from March 2024 to March 31, 2025, and CFO Health at Reckitt from April 2021
to February 2024. In his EVP Corporate Controller role with Reckitt, Mr. Bealer led a global team of over 600 employees accountable
for global controls, financial shared services, group reporting and accounting, as well as leading the functional operating model transformation.
Prior to his roles at Reckitt, M
[... clipped after ...]
\end{trajectorylisting}

\begin{trajectorylisting}{toolNeutral}{State after turn 4}
pool_size: 7
curated_size: 7/30
importance: very_high=0, high=3, fair=3, low=1
top_curated:
  - 0001104659-25-006698_0 [high]
  - 0001104659-25-053287_0 [high]
  - 0001104659-25-121934_20 [high]
  - 0000909832-25-000101_19 [fair]
  - 0001364954-25-000039_57 [fair]
  - 0001673985-25-000053_24 [fair]
  - 0001530950-25-000260_18 [low]
\end{trajectorylisting}

\subsubsection*{\textcolor{toolGrep!80!black}{Phase 3 - Exact grep adds corroborating filings}}
\begin{trajectorylisting}{toolGrep}{Exact grep call}
[
  {
    "tool": "grep_corpus",
    "params": {
      "pattern": "effective July 3, 2025"
    }
  }
]
\end{trajectorylisting}

\begin{trajectorylisting}{toolGrep}{Grep return}
[... clipped before ...]
| CHRISTOPHER J.          
 BEALER                  |  |  |  | TIMOTHY R.                                    
 KRAFT                                         |  |  |  | AMY C.                                
 HELD                                  |  |  |  | RYAN A.                 
 THOMAS                  |  |  |  | SHAUN P.                       
 MARA                           |  |
|  |  | President and Chief Executive Officer |  |  |  | Chief Financial Officer |  |  |  | Chief Legal and Corporate Development Officer |  |  |  | SVP and Chief Human Resources Officer |  |  |  | SVP and General         
 Manager Atkins and OWYN |  |  |  | Former Chief Financial Officer |  |

Mr. Bealer became an NEO for the first time in fiscal year 2025. Mr. Bealer was hired on March 31, 2025. Mr. Mara retired as our Chief Financial Officer and resigned effective July 3, 2025. Mr. Mara's executive compensation is outlined further in the compensation tables beginning on page 54 but is otherwise not discussed with other NEOs except where noted. Executive Summary Our compensation program for our NEOs and other executive officers is designed to meet the following primary objectives:

|  |  | Management Development and Continuity.                                                                                                                                                                                                   
 Attract, retain and motivate individuals of superior ability and managerial talent to develop, grow and manage our business by offering competitive compensation opportunities with both short-term and significant long-term components |  |  |  | Pay for Performance.
[... clipped after ...]
\end{trajectorylisting}

\begin{trajectorylisting}{toolCurate}{Curation update}
Curated set updated (9/30): 0001104659-25-006698_0[high], 0001104659-25-053287_0[high], 0001104659-25-121934_20[high], 0001104659-25-121934_21[high], 0001104659-25-121934_42[high], 0000909832-25-000101_19[fair], 0001364954-25-000039_57[fair], 0001673985-25-000053_24[fair], 0001530950-25-000260_18[low]
\end{trajectorylisting}

\begin{trajectorylisting}{toolNeutral}{State after turn 7}
pool_size: 10
curated_size: 9/30
importance: very_high=0, high=5, fair=3, low=1
top_curated:
  - 0001104659-25-006698_0 [high]
  - 0001104659-25-053287_0 [high]
  - 0001104659-25-121934_20 [high]
  - 0001104659-25-121934_21 [high]
  - 0001104659-25-121934_42 [high]
  - 0000909832-25-000101_19 [fair]
  - 0001364954-25-000039_57 [fair]
  - 0001673985-25-000053_24 [fair]
  - 0001530950-25-000260_18 [low]
\end{trajectorylisting}

\clearpage
\subsection{Case Study B: Fan-Out Search Resolves an Ambiguous Transfer Query}
\label{app:case-frames461}

\begin{promptbox}{FRAMES query}
\textbf{Query.} Which university did the actor who has appeared in the most movies in the American pie film series (including spinoffs, as of 2020) deliver a commencement address at in 2012?\\[2pt]
\textbf{Gold answer.} Dalhousie University, in Halifax, Nova Scotia\\[2pt]
\textbf{Outcome.} 16 turns, tool diversity $6$, curated recall $1.0$, final-answer recall $1.0$.
\end{promptbox}

\noindent\textbf{Why this case is interesting.}
The query is ambiguous at the first hop: several \emph{American Pie} actors are plausible, and the commencement-address clue only becomes useful after the actor identity is narrowed. The trajectory shows why fan-out and tagged memory matter. The agent keeps competing actor hypotheses alive, pivots from Seann William Scott to Eugene Levy, then uses exact search and full-document reading to expose the Dalhousie address evidence.

\noindent\textbf{How the trajectory is abridged.}
The full log includes many generic web snippets from fan-out search. Here we omit generic pages, repeated low-impact curation text, and the no-op verifier call, while preserving the reasoning and exact calls for the decisive fan-out, candidate pivot, exact search, full read, and final promotion turns.

\paragraph{Trajectory overview.}
{\small
\setlength{\tabcolsep}{3pt}
\renewcommand{\arraystretch}{1.08}
\begin{longtable}{p{0.055\textwidth}p{0.12\textwidth}p{0.36\textwidth}p{0.36\textwidth}}
\toprule
Turn & Tool & Policy move & Harness state / why it matters \\
\midrule
\endfirsthead
\toprule
Turn & Tool & Policy move & Harness state / why it matters \\
\midrule
\endhead
0 & \CaseTool{toolFanout}{fan-out} & Fan out over actor identity and address clues & auto-seed stores 8 docs; broader pool contains the key American Pie character page \\
1 & \CaseTool{toolCurate}{curate} & Promote American Pie cast evidence & curation shifts memory from generic hits to candidate actors \\
2 & \CaseTool{toolSearch}{search} & Search Eugene Levy commencement & tests the candidate implied by Noah Levenstein \\
3 & \CaseTool{toolCurate}{curate} & Add Eugene Levy pages & memory now tracks the alternative to Seann William Scott \\
4 & \CaseTool{toolSearch}{search} & Search most-appearances clue & keeps the actor-identity subproblem explicit \\
5 & \CaseTool{toolCurate}{curate} & Promote cast pages & curated set expands but stays tagged \\
6 & \CaseTool{toolSearch}{search} & Search Seann William Scott address & tests a competing hypothesis \\
7 & \CaseTool{toolCurate}{curate} & Demote weak address hits & low tags protect the final set from noise \\
8 & \CaseTool{toolGrep}{grep} & Exact grep for both candidates & fails to settle the address but broadens evidence \\
9 & \CaseTool{toolCurate}{curate} & Fill remaining memory slots & curated set reaches cap; tags govern later eviction \\
12 & \CaseTool{toolSearch}{search} & Search Eugene Levy + 2012 address & directly targets the missing bridge \\
13 & \CaseTool{toolCurate}{curate} & Add university/address candidates & state prepares for final read \\
14 & \CaseTool{toolRead}{read} & Read Eugene Levy page & full text reveals Dalhousie University \\
15 & \CaseTool{toolCurate}{curate} & Promote Eugene Levy and Dalhousie & final answer evidence receives very\_high/high tags \\
\bottomrule
\end{longtable}
}
\subsubsection*{\textcolor{toolFanout!80!black}{Phase 1 - Fan-out keeps competing hypotheses alive}}
\begin{trajectorylisting}{toolNeutral}{Reasoning excerpt}
We need: Which university did the actor who has appeared in the most movies in the American Pie film series (including spinoffs, as of 2020) deliver a commencement address at in 2012.

We can break it down.

Candidates: Need actor with most appearances in American Pie series.

List of movies: American Pie (1999), American Pie 2 (2001), American Pie 3 (2003), American Pie 4 (2008), American Wedding (2003?), Wait 2003? Actually American Wedding (2003) is the third? Let's check full series: American Pie (1999); American Pie 2 (2001); American Wedding (2003); American Reunion (2012); American Pie Presents: Band Party (2007); American Pie Presents: The Wedding (2009); American Pie Presents: The Book of Love (2010); American Pie Presents: The Naked Mile (2006); American Pie Presents: The First Kiss (2009). Also spinoffs: American Wedding (2003), American Reunion 2012. The actor who appears in the most movies: I recall Skyler or others? Let's recall main casts: Jason Biggs (Jim), Alyson Hannigan (Michelle), Natasha Lyonne (Stifler?), actually Stifler is Seann William Scott. He appears in many? Let's count: Stifler from first through fourth, plus spinoffs? Seann William Scott appears in 1) American Pie 2? Wait Stifler is in first: yes first. In American Pie 2? yes Stifler. In American Wedding? maybe he appears as Stifler. In American Pie Presents: The Book of Love? Not sure. Also spinoffs: He appears in American Pie Presents: The Book of Love? Let's recall: Seann William Scott as Stifler appears in original series plus the spin-offs? There's also his film within the series. Let's search.

So likely the actor is Seann William Scott, or maybe the actor is Tom Gills? No. The actor with most appearances likely is Seann William Scott or Jason Biggs? Jason Biggs appears first through first three? He appears also in reunion and maybe New movies. Let's verify counts.

We need actor with most appearances among all US movies up to 2020. Let's search for 'Seann William Scott American Pie appearances' or 'American Pie actor with most appearances'.

We also need commencement address in 2012 at a university. Which university? So Seann William Scott delivered commencement address in 2012 at some university. Need to find that.

So we need to find doc about Seann William Scott commencement address 2012. Alternatively, if actor is Jason Biggs, find if he delivered commencement address 2012. But Jason Biggs may have spoken at University of Notre Dame? Let's think: Jason Biggs Speech 2013? Wait.

Let's recall: Seann William Scott on his biography: He gave a commencement address at University of Kansas? Let's quickly search.

Given time, best to do fan_out_search queries.

First query: "Seann William Scott commencement address 2012 university".

Second query: "American Pie series actor most appearances".

Third: "American Pie spinoffs actor who appears in most films".

Fourth: "Jason Biggs commencement address 2012".

Fifth: "American Pie actor commencement 2012".

But my first size: "Seann William Scott commencement address 2012 university" should find doc. Let's do that.
\end{trajectorylisting}

\begin{trajectorylisting}{toolFanout}{Exact fan-out call}
[
  {
    "tool": "fan_out_search",
    "params": {
      "queries": [
        "Seann William Scott commencement address 2012 university",
        "most American Pie films actor",
        "American Pie actor who appeared most movies",
        "Seann William Scott 2012 commencement",
        "Jason Biggs commencement 2012"
      ]
    }
  }
]
\end{trajectorylisting}

\begin{trajectorylisting}{toolFanout}{Fan-out return: actor identity clue}
[... clipped before ...]
cott - Wikipedia
Scott is best known for his breakout role as Steve Stifler in the American Pie film series (1999-2012). He also starred in a lead role as Doug Glatt in Goon ( ...

# DOCUMENT ID: https://en.wikipedia.org/wiki/Thomas_Ian_Nicholas (44 tokens)
Thomas Ian Nicholas - Wikipedia
In film, Nicholas is best known for his lead roles as Kevin Meyers in the American Pie theatrical film series (1999-2012) and Henry Rowengartner in the sports ...

# DOCUMENT ID: https://en.wikipedia.org/wiki/Chris_Owen_(actor) (46 tokens)
Chris Owen (actor) - Wikipedia
Chris Owen (born September 25, 1980) is an American actor. He is best known for his supporting role as Chuck "Sherminator" Sherman in the American Pie ...

# DOCUMENT ID: https://en.wikipedia.org/wiki/List_of_American_Pie_characters (50 tokens)
List of American Pie characters - Wikipedia
Noah Levenstein, played by Eugene Levy, has appeared in eight of the nine released films. Biggs, Hannigan, Scott, Nicholas, Thomas, Coolidge, Cho, Isfield, ...

# DOCUMENT ID: https://en.wikipedia.org/wiki/Tad_Hilgenbrink (52 tokens)
Tad Hilgenbrink - Wikipedia
Tad Hilgenbrink (born October 9, 1981) is an American actor, known for his role as Matt Stifler in American Pie Presents: Band Camp, Chance Marquis in The ...

# DOCUMENT ID: https://en.wikipedia.org/wiki/Chris_Klein_(actor) (49 tokens)
Chris Klein (actor) - Wikipedia
He reprised his role in two sequels: American Pie 2 (2001) and American Reunion (2012). Chris Klein. Klein in March 2012. Born. Frederick Christopher Klein.

# DOCUMENT ID: https://en.wikipedia.org/wiki/Eddie_Kaye_Thomas (41 tokens)
Eddie Kaye Thomas - Wikipedia
Thomas rose to prominence for his supporting role as Pau
[... clipped after ...]
\end{trajectorylisting}

\begin{trajectorylisting}{toolCurate}{Curation update}
Curated set updated (15/30): https://en.wikipedia.org/wiki/American_Pie_(film_series)[high], https://en.wikipedia.org/wiki/List_of_American_Pie_characters[high], https://en.wikipedia.org/wiki/Seann_William_Scott[high], https://en.wikipedia.org/wiki/Ben_Carson[fair], https://en.wikipedia.org/wiki/Bradley_Cooper[fair], https://en.wikipedia.org/wiki/Chris_Owen_(actor)[fair], https://en.wikipedia.org/wiki/Ellis_Paul[fair], https://en.wikipedia.org/wiki/Jason_Biggs[fair], https://en.wikipedia.org/wiki/Magic_Johnson[fair], https://en.wikipedia.org/wiki/Steven_Curtis_Chapman[fair], https://en.wikipedia.org/wiki/Susan_Rice[fair], https://en.wikipedia.org/wiki/Thomas_Ian_Nicholas[fair], https://en.wikipedia.org/wiki/Tim_Reid[fair], https://en.wikipedia.org/wiki/Tucker_Carlson[fair], https://en.wikipedia.org/wiki/Chris_Klein_(actor)[low]
\end{trajectorylisting}

\begin{trajectorylisting}{toolNeutral}{State after turn 1}
pool_size: 40
curated_size: 15/30
importance: very_high=0, high=3, fair=11, low=1
top_curated:
  - https://en.wikipedia.org/wiki/American_Pie_(film_series) [high]
  - https://en.wikipedia.org/wiki/List_of_American_Pie_characters [high]
  - https://en.wikipedia.org/wiki/Seann_William_Scott [high]
  - https://en.wikipedia.org/wiki/Ben_Carson [fair]
  - https://en.wikipedia.org/wiki/Bradley_Cooper [fair]
  - https://en.wikipedia.org/wiki/Chris_Owen_(actor) [fair]
  - https://en.wikipedia.org/wiki/Ellis_Paul [fair]
  - https://en.wikipedia.org/wiki/Jason_Biggs [fair]
  - https://en.wikipedia.org/wiki/Magic_Johnson [fair]
  - https://en.wikipedia.org/wiki/Steven_Curtis_Chapman [fair]
  - https://en.wikipedia.org/wiki/Susan_Rice [fair]
  - https://en.wikipedia.org/wiki/Thomas_Ian_Nicholas [fair]
  ... 3 more
\end{trajectorylisting}

\subsubsection*{\textcolor{toolSearch!80!black}{Phase 2 - The policy pivots from Seann William Scott to Eugene Levy}}
\begin{trajectorylisting}{toolSearch}{Search Eugene Levy address}
[
  {
    "tool": "search_corpus",
    "params": {
      "query": "Eugene Levy 2012 commencement address university"
    }
  }
]
\end{trajectorylisting}

\begin{trajectorylisting}{toolSearch}{Search most-appearances clue}
[
  {
    "tool": "search_corpus",
    "params": {
      "query": "most American Pie films actor appeared in most movies american pie series"
    }
  }
]
\end{trajectorylisting}

\begin{trajectorylisting}{toolCurate}{Curation after candidate expansion}
Curated set updated (21/30): https://en.wikipedia.org/wiki/American_Pie_(film)[high], https://en.wikipedia.org/wiki/American_Pie_(film_series)[high], https://en.wikipedia.org/wiki/Eugene_Levy[high], https://en.wikipedia.org/wiki/List_of_American_Pie_characters[high], https://en.wikipedia.org/wiki/Seann_William_Scott[high], https://en.wikipedia.org/wiki/Ben_Carson[fair], https://en.wikipedia.org/wiki/Bradley_Cooper[fair], https://en.wikipedia.org/wiki/Chris_Owen_(actor)[fair], https://en.wikipedia.org/wiki/Ellis_Paul[fair], https://en.wikipedia.org/wiki/Jason_Biggs[fair], https://en.wikipedia.org/wiki/John_Cho[fair], https://en.wikipedia.org/wiki/Magic_Johnson[fair], https://en.wikipedia.org/wiki/Shannon_Elizabeth[fair], https://en.wikipedia.org/wiki/Steven_Curtis_Chapman[fair], https://en.wikipedia.org/wiki/Susan_Rice[fair], https://en.wikipedia.org/wiki/Thomas_Ian_Nicholas[fair], https://en.wikipedia.org/wiki/Tim_Reid[fair], https://en.wikipedia.org/wiki/Tucker_Carlson[fair], https://en.wikipedia.org/wiki/Chris_Klein_(actor)[low], https://en.wikipedia.org/wiki/Dan_Petronijevic[low], https://en.wikipedia.org/wiki/Eugene_Levy_(politician)[low]
\end{trajectorylisting}

\begin{trajectorylisting}{toolNeutral}{State after turn 5}
pool_size: 53
curated_size: 21/30
importance: very_high=0, high=5, fair=13, low=3
top_curated:
  - https://en.wikipedia.org/wiki/American_Pie_(film) [high]
  - https://en.wikipedia.org/wiki/American_Pie_(film_series) [high]
  - https://en.wikipedia.org/wiki/Eugene_Levy [high]
  - https://en.wikipedia.org/wiki/List_of_American_Pie_characters [high]
  - https://en.wikipedia.org/wiki/Seann_William_Scott [high]
  - https://en.wikipedia.org/wiki/Ben_Carson [fair]
  - https://en.wikipedia.org/wiki/Bradley_Cooper [fair]
  - https://en.wikipedia.org/wiki/Chris_Owen_(actor) [fair]
  - https://en.wikipedia.org/wiki/Ellis_Paul [fair]
  - https://en.wikipedia.org/wiki/Jason_Biggs [fair]
  - https://en.wikipedia.org/wiki/John_Cho [fair]
  - https://en.wikipedia.org/wiki/Magic_Johnson [fair]
  ... 9 more
\end{trajectorylisting}

\subsubsection*{\textcolor{toolGrep!80!black}{Phase 3 - Exact search tests the candidate bridge}}
\begin{trajectorylisting}{toolGrep}{Exact grep call}
[
  {
    "tool": "grep_corpus",
    "params": {
      "pattern": "Seann William Scott commencement|Eugene Levy commencement|Seann William Scott address|Eugene Levy address|2012 commencement"
    }
  }
]
\end{trajectorylisting}

\begin{trajectorylisting}{toolGrep}{Grep return excerpt}
# DOCUMENT ID: https://en.wikipedia.org/wiki/Donald_Sutherland (44 tokens) Donald Sutherland - Wikipedia Donald McNichol Sutherland (17 July 1935 - 20 June 2024) was a Canadian actor. With a career spanning six decades, he received numerous accolades, ...

# DOCUMENT ID: https://en.wikipedia.org/wiki/John_Lithgow (36 tokens) John Lithgow - Wikipedia John Arthur Lithgow is an American actor. He studied at Harvard University and the London Academy of Music and Dramatic Art before becoming known for his ...
\end{trajectorylisting}

\begin{trajectorylisting}{toolSearch}{Direct bridge search}
[
  {
    "tool": "search_corpus",
    "params": {
      "query": "Eugene Levy 2012 commencement address university biographical speaker 2012"
    }
  }
]
\end{trajectorylisting}

\begin{trajectorylisting}{toolSearch}{Search return before full read}
# DOCUMENT ID: https://en.wikipedia.org/wiki/Princeton_University (38 tokens) Princeton University - Wikipedia Princeton University is a private Ivy League research university in Princeton, New Jersey, United States. Founded in 1746 in Elizabeth as the College of New ... # DOCUMENT ID: https://en.wikipedia.org/wiki/List_of_University_of_Michigan_alumni (41 tokens) List of University of Michigan alumni - Wikipedia The following is a list of University of Michigan alumni. There are more than 640,000 living alumni of the University of Michigan in 180 countries across ... # DOCUMENT ID: https://en.wikipedia.org/wiki/List_of_Cornell_University_alumni (31 tokens) List of Cornell University alumni - Wikipedia This list of Cornell University alumni includes notable graduates, non-graduate former students, and current students of Cornell University. # DOCUMENT ID: https://en.wikipedia.org/wiki/University_of_St._Thomas_(Minnesota) (37 tokens) University of St. Thomas (Minnesota) - Wikipedia The University of St. Thomas is a private Catholic research university with campuses in Saint Paul and Minneapolis, Minnesota, United States. # DOCUMENT ID: https://en.wikipedia.org/wiki/List_of_Queens_College_people (43 tokens) List of Queens College people - Wikipedia This is a list of notable alumni and faculty of Queens College, City University of New York. Contents. 1 Business; 2 Education; 3 ...
\end{trajectorylisting}

\begin{trajectorylisting}{toolNeutral}{State after turn 13}
pool_size: 74
curated_size: 30/30
importance: very_high=0, high=5, fair=14, low=11
top_curated:
  - https://en.wikipedia.org/wiki/American_Pie_(film) [high]
  - https://en.wikipedia.org/wiki/American_Pie_(film_series) [high]
  - https://en.wikipedia.org/wiki/Eugene_Levy [high]
  - https://en.wikipedia.org/wiki/List_of_American_Pie_characters [high]
  - https://en.wikipedia.org/wiki/Seann_William_Scott [high]
  - https://en.wikipedia.org/wiki/Ben_Carson [fair]
  - https://en.wikipedia.org/wiki/Bradley_Cooper [fair]
  - https://en.wikipedia.org/wiki/Chris_Owen_(actor) [fair]
  - https://en.wikipedia.org/wiki/Ellis_Paul [fair]
  - https://en.wikipedia.org/wiki/Jason_Biggs [fair]
  - https://en.wikipedia.org/wiki/John_Cho [fair]
  - https://en.wikipedia.org/wiki/Magic_Johnson [fair]
  ... 18 more
\end{trajectorylisting}

\subsubsection*{\textcolor{toolRead!80!black}{Phase 4 - Full-document reading exposes the answer sentence}}
\begin{trajectorylisting}{toolRead}{Exact read call}
[
  {
    "tool": "read_document",
    "params": {
      "doc_id": "https://en.wikipedia.org/wiki/Eugene_Levy"
    }
  }
]
\end{trajectorylisting}

\begin{trajectorylisting}{toolRead}{Full-document excerpt around answer}
[... clipped before ...]
me honourees, alongside fellow SCTV alumni John Candy, Martin Short, and Catherine O'Hara. Levy is one of only a handful of people who have won at least five Canadian Comedy Awards, including two for Best Writing (Best in Show in 2001 and A Mighty Wind in 2004) and three for Best Male Performer (Best in Show, American Pie 2 in 2002, and A Mighty Wind).
In 2008, the governor general of Canada presented Levy with the Governor General's Performing Arts Awards (GGPAA), a lifetime achievement award considered "for their outstanding body of work and enduring contribution to the performing arts in Canada.". In 2010, Levy was awarded the ACTRA Award by the union representing Canada's actors. In 2011, Levy was made a Member of the Order of Canada "for his contributions as a comic actor and writer, and for his dedication to charitable causes" and promoted to the rank of Companion in 2022.
On May 22, 2012, Levy delivered a commencement address at Dalhousie University, in Halifax, Nova Scotia, and was awarded the degree Doctor of Laws (honoris causa). On June 11, 2012, Levy was presented with the Queen Elizabeth II Diamond Jubilee Medal by the lieutenant governor of Ontario.
On March 13, 2016, Levy took home the award for "Best Performance by an Actor in a Continuing Leading Comedic Role" at the 4th Annual Canadian Screen Awards, for his performance as Johnny Rose in the CBC/Pop TV sitcom Schitt's Creek.

References
External links

Eugene Levy at IMDb
Eugene Levy at the Internet Broadway Database
Eugene Levy discography at Discogs
Article at thecanadianencyclopedia.ca
Eugene Levy on X
[Context: 9839/30720]
\end{trajectorylisting}

\begin{trajectorylisting}{toolCurate}{Final curation update}
Curated set updated (30/30): https://en.wikipedia.org/wiki/Eugene_Levy[very_high], https://en.wikipedia.org/wiki/American_Pie_(film)[high], https://en.wikipedia.org/wiki/American_Pie_(film_series)[high], https://en.wikipedia.org/wiki/Dalhousie_University[high], https://en.wikipedia.org/wiki/List_of_American_Pie_characters[high], https://en.wikipedia.org/wiki/Seann_William_Scott[high], https://en.wikipedia.org/wiki/Ben_Carson[fair], https://en.wikipedia.org/wiki/Bradley_Cooper[fair], https://en.wikipedia.org/wiki/Chris_Owen_(actor)[fair], https://en.wikipedia.org/wiki/Ellis_Paul[fair], https://en.wikipedia.org/wiki/Jason_Biggs[fair], https://en.wikipedia.org/wiki/John_Cho[fair], https://en.wikipedia.org/wiki/Magic_Johnson[fair], https://en.wikipedia.org/wiki/Princeton_University[fair], https://en.wikipedia.org/wiki/Shannon_Elizabeth[fair], https://en.wikipedia.org/wiki/Steven_Curtis_Chapman[fair], https://en.wikipedia.org/wiki/Susan_Rice[fair], https://en.wikipedia.org/wiki/Thomas_Ian_Nicholas[fair], https://en.wikipedia.org/wiki/Tim_Reid[fair], https://en.wikipedia.org/wiki/Tucker_Carlson[fair], https://en.wikipedia.org/wiki/Bradley_Whitford[low], https://en.wikipedia.org/wiki/Dan_Petronijevic[low], https://en.wikipedia.org/wiki/David_Sutherland[low], https://en.wikipedia.org/wiki/Donald_Sutherland[low], https://en.wikipedia.org/wiki/Heath_Ledger[low], https://en.wikipedia.org/wiki/John_Lithgow[low], https://en.wikipedia.org/wiki/Jonathan_Franzen[low], https://en.wikipedia.org/wiki/Joss_Whedon[low], https://en.wikipedia.org/wiki/List_of_NYU_Tisch_School_of_the_Arts_people[low], https://en.wikipedia.org/wiki/Wes_Craven[low]
[EVICTED low-importance] 1 doc(s): https://en.wikipedia.org/wiki/Eugene_Levy_(politician)
\end{trajectorylisting}

\begin{trajectorylisting}{toolNeutral}{State after turn 15}
pool_size: 74
curated_size: 30/30
importance: very_high=1, high=5, fair=14, low=10
top_curated:
  - https://en.wikipedia.org/wiki/Eugene_Levy [very_high]
  - https://en.wikipedia.org/wiki/American_Pie_(film) [high]
  - https://en.wikipedia.org/wiki/American_Pie_(film_series) [high]
  - https://en.wikipedia.org/wiki/Dalhousie_University [high]
  - https://en.wikipedia.org/wiki/List_of_American_Pie_characters [high]
  - https://en.wikipedia.org/wiki/Seann_William_Scott [high]
  - https://en.wikipedia.org/wiki/Ben_Carson [fair]
  - https://en.wikipedia.org/wiki/Bradley_Cooper [fair]
  - https://en.wikipedia.org/wiki/Chris_Owen_(actor) [fair]
  - https://en.wikipedia.org/wiki/Ellis_Paul [fair]
  - https://en.wikipedia.org/wiki/Jason_Biggs [fair]
  - https://en.wikipedia.org/wiki/John_Cho [fair]
  ... 18 more
\end{trajectorylisting}

\clearpage
\subsection{Case Study D: Held-Out Multi-Hop QA via Search-State Editing}
\label{app:case-hotpot-snl}

\begin{promptbox}{HotpotQA query}
\textbf{Query.} Which American comedian starred in the original cast of Saturday Night Live and played Stan Winters in the sitcom Martin?\\[2pt]
\textbf{Gold answer.} Garrett Morris\\[2pt]
\textbf{Outcome.} 11 turns, tool diversity $5$, curated recall $1.0$, final-answer recall $1.0$.
\end{promptbox}

\noindent\textbf{Why this case matters.}
This held-out HotpotQA example is intentionally small: it makes the transfer claim visible at trajectory level. The policy applies the same operations learned elsewhere, namely search, curation, full-document reading, exact grep, and final evidence promotion, to solve a multi-hop web query outside the SFT and RL families.

\paragraph{Trajectory overview.}
{\small
\setlength{\tabcolsep}{3pt}
\renewcommand{\arraystretch}{1.08}
\begin{longtable}{p{0.055\textwidth}p{0.12\textwidth}p{0.36\textwidth}p{0.36\textwidth}}
\toprule
Turn & Tool & Policy move & Harness state / why it matters \\
\midrule
\endfirsthead
\toprule
Turn & Tool & Policy move & Harness state / why it matters \\
\midrule
\endhead
0 & \CaseTool{toolSearch}{search} & Search both clues jointly & initial hit identifies Garrett Morris through the Stan Winters clue \\
1 & \CaseTool{toolCurate}{curate} & Promote candidate pages & auto-seeded web hits become an editable curated set \\
2 & \CaseTool{toolSearch}{search} & Search the SNL-original-cast clue & second hop confirms the other half of the query \\
3 & \CaseTool{toolCurate}{curate} & Promote SNL cast evidence & curated state now contains both supporting paths \\
4 & \CaseTool{toolRead}{read} & Read Garrett Morris page & full text unifies comedian, original SNL cast, and Martin role \\
5 & \CaseTool{toolCurate}{curate} & Keep Garrett Morris high & final answer evidence is retained \\
6 & \CaseTool{toolGrep}{grep} & Grep exact role phrase & exact matching checks the Stan Winters clue \\
7 & \CaseTool{toolCurate}{curate} & Add sitcom/role pages & curated set adds corroborating context \\
\bottomrule
\end{longtable}
}
\subsubsection*{\textcolor{toolSearch!80!black}{Phase 1 - Joint search identifies the likely entity}}
\begin{trajectorylisting}{toolSearch}{Exact search call}
[
  {
    "tool": "search_corpus",
    "params": {
      "query": "Stan Winters Martin SNL original cast comedian"
    }
  }
]
\end{trajectorylisting}

\begin{trajectorylisting}{toolSearch}{Search return: Stan Winters points to Garrett Morris}
# DOCUMENT ID: https://en.wikipedia.org/wiki/Garrett_Morris (46 tokens) Garrett Morris - Wikipedia He is also known for his role in the sitcom Martin as Stan Winters, from 1992 to 1995, until he suffered an injury. Also, he made two guest appearances on The ...
\end{trajectorylisting}

\begin{trajectorylisting}{toolCurate}{Curation update}
Curated set updated (8/30): https://en.wikipedia.org/wiki/Garrett_Morris[high], https://en.wikipedia.org/wiki/Andrea_Martin[fair], https://en.wikipedia.org/wiki/Bobby_Moynihan[fair], https://en.wikipedia.org/wiki/Don_Novello[fair], https://en.wikipedia.org/wiki/Richard_Lewis_(comedian)[fair], https://en.wikipedia.org/wiki/Second_City_Television[fair], https://en.wikipedia.org/wiki/Victoria_Jackson[fair], https://en.wikipedia.org/wiki/Will_Forte[fair]
\end{trajectorylisting}

\begin{trajectorylisting}{toolNeutral}{State after turn 1}
pool_size: 10
curated_size: 8/30
importance: very_high=0, high=1, fair=7, low=0
top_curated:
  - https://en.wikipedia.org/wiki/Garrett_Morris [high]
  - https://en.wikipedia.org/wiki/Andrea_Martin [fair]
  - https://en.wikipedia.org/wiki/Bobby_Moynihan [fair]
  - https://en.wikipedia.org/wiki/Don_Novello [fair]
  - https://en.wikipedia.org/wiki/Richard_Lewis_(comedian) [fair]
  - https://en.wikipedia.org/wiki/Second_City_Television [fair]
  - https://en.wikipedia.org/wiki/Victoria_Jackson [fair]
  - https://en.wikipedia.org/wiki/Will_Forte [fair]
\end{trajectorylisting}

\subsubsection*{\textcolor{toolSearch!80!black}{Phase 2 - Second-hop search checks original SNL cast membership}}
\begin{trajectorylisting}{toolSearch}{Exact search call}
[
  {
    "tool": "search_corpus",
    "params": {
      "query": "Garrett Morris Saturday Night Live original cast"
    }
  }
]
\end{trajectorylisting}

\begin{trajectorylisting}{toolSearch}{Search return: original cast evidence}
# DOCUMENT ID: https://en.wikipedia.org/wiki/Saturday_Night_Live_15th_Anniversary_Special (46 tokens)
Saturday Night Live 15th Anniversary Special - Wikipedia
The special was SNL's first anniversary celebration; it featured original cast members Chevy Chase, Dan Aykroyd, Jane Curtin, Garrett Morris, and Laraine ...

# DOCUMENT ID: https://en.wikipedia.org/wiki/List_of_Saturday_Night_Live_cast_members (54 tokens)
List of Saturday Night Live cast members - Wikipedia
Original cast (from left to right): Laraine Newman, John Belushi, Jane Curtin, Gilda Radner, Dan Aykroyd, Garrett Morris, and Chevy Chase. Contents. 1 List.

# DOCUMENT ID: https://en.wikipedia.org/wiki/Saturday_Night_Live_season_1 (56 tokens)
Saturday Night Live season 1 - Wikipedia
Starring. Dan Aykroyd * John Belushi * Chevy Chase * George Coe * Jane Curtin * Garrett Morris * Laraine Newman * Michael O'Donoghue * Gilda Radner ; No. of ...

# DOCUMENT ID: https://en.wikipedia.org/wiki/Saturday_Night_Live_season_3 (40 tokens)
Saturday Night Live season 3 - Wikipedia
Cast * Dan Aykroyd * John Belus
[... clipped after ...]
\end{trajectorylisting}

\begin{trajectorylisting}{toolCurate}{Curation update}
Curated set updated (12/30): https://en.wikipedia.org/wiki/Garrett_Morris[high], https://en.wikipedia.org/wiki/List_of_Saturday_Night_Live_cast_members[high], https://en.wikipedia.org/wiki/Saturday_Night_Live_15th_Anniversary_Special[high], https://en.wikipedia.org/wiki/Andrea_Martin[fair], https://en.wikipedia.org/wiki/Bobby_Moynihan[fair], https://en.wikipedia.org/wiki/Don_Novello[fair], https://en.wikipedia.org/wiki/Richard_Lewis_(comedian)[fair], https://en.wikipedia.org/wiki/Saturday_Night_Live_season_1[fair], https://en.wikipedia.org/wiki/Saturday_Night_Live_season_3[fair], https://en.wikipedia.org/wiki/Second_City_Television[fair], https://en.wikipedia.org/wiki/Victoria_Jackson[fair], https://en.wikipedia.org/wiki/Will_Forte[fair]
\end{trajectorylisting}

\begin{trajectorylisting}{toolNeutral}{State after turn 3}
pool_size: 19
curated_size: 12/30
importance: very_high=0, high=3, fair=9, low=0
top_curated:
  - https://en.wikipedia.org/wiki/Garrett_Morris [high]
  - https://en.wikipedia.org/wiki/List_of_Saturday_Night_Live_cast_members [high]
  - https://en.wikipedia.org/wiki/Saturday_Night_Live_15th_Anniversary_Special [high]
  - https://en.wikipedia.org/wiki/Andrea_Martin [fair]
  - https://en.wikipedia.org/wiki/Bobby_Moynihan [fair]
  - https://en.wikipedia.org/wiki/Don_Novello [fair]
  - https://en.wikipedia.org/wiki/Richard_Lewis_(comedian) [fair]
  - https://en.wikipedia.org/wiki/Saturday_Night_Live_season_1 [fair]
  - https://en.wikipedia.org/wiki/Saturday_Night_Live_season_3 [fair]
  - https://en.wikipedia.org/wiki/Second_City_Television [fair]
  - https://en.wikipedia.org/wiki/Victoria_Jackson [fair]
  - https://en.wikipedia.org/wiki/Will_Forte [fair]
\end{trajectorylisting}

\subsubsection*{\textcolor{toolRead!80!black}{Phase 3 - Full-document reading unifies both clues}}
\begin{trajectorylisting}{toolRead}{Exact read call}
[
  {
    "tool": "read_document",
    "params": {
      "doc_id": "https://en.wikipedia.org/wiki/Garrett_Morris"
    }
  }
]
\end{trajectorylisting}

\begin{trajectorylisting}{toolRead}{Full-document excerpt around answer}
# Document (2665 tokens)
# Garrett Morris

## Summary
Garrett Isaac Morris (born February 1, 1937) is an American actor, comedian and singer. He was part of the original cast and was the first Black cast member of the sketch comedy program Saturday Night Live, appearing from 1975 to 1980.
He also played Jimmy on The Jeffersons (1983-1984). Morris had one of the starring roles, as Junior "Uncle Junior" King, on the sitcom The Jamie Foxx Show, which aired from 1996 to 2001. Morris also had a regular role as Earl Washington on the CBS sitcom 2 Broke Girls, from 2011 to 2017.
He is also known for his role in the sitcom Martin as Stan Winters, from 1992 to 1995, until he suffered an injury. Also, he made two guest appearances on The Wayans Bros. in season one, episode one as himself and again on episode ten as the brothers' uncle Leon (1995). He played a concerned teacher in the film Cooley High (1975), Slide in Car Wash (1976), and Carl in The Census Taker (1984).
In 2024, Morris was honored with his own star on the Hollywood Walk of Fame.

## Full Text
Garrett Isaac Morris (born February 1, 1937)
[... clipped after ...]
\end{trajectorylisting}

\begin{trajectorylisting}{toolGrep}{Exact grep call}
[
  {
    "tool": "grep_corpus",
    "params": {
      "pattern": "Stan Winters"
    }
  }
]
\end{trajectorylisting}

\begin{trajectorylisting}{toolGrep}{Grep return}
# DOCUMENT ID: https://en.wikipedia.org/wiki/Garrett_Morris (45 tokens) Garrett Morris - Wikipedia He is also known for his role in the sitcom Martin as Stan Winters, from 1992 to 1995, until he suffered an injury. Also, he made two guest appearances on ...

# DOCUMENT ID: https://en.wikipedia.org/wiki/Martin_(TV_series) (55 tokens) Martin (TV series) - Wikipedia Garrett Morris as Stan Winters (seasons 1-2; guest season 3). Supporting characters. edit * Reginald Ballard as Bruh-Man (Brother Man) (seasons 2-4); Tracy ...
\end{trajectorylisting}

\begin{trajectorylisting}{toolNeutral}{State after turn 7}
pool_size: 23
curated_size: 14/30
importance: very_high=0, high=3, fair=10, low=1
top_curated:
  - https://en.wikipedia.org/wiki/Garrett_Morris [high]
  - https://en.wikipedia.org/wiki/List_of_Saturday_Night_Live_cast_members [high]
  - https://en.wikipedia.org/wiki/Saturday_Night_Live_15th_Anniversary_Special [high]
  - https://en.wikipedia.org/wiki/Andrea_Martin [fair]
  - https://en.wikipedia.org/wiki/Bobby_Moynihan [fair]
  - https://en.wikipedia.org/wiki/Don_Novello [fair]
  - https://en.wikipedia.org/wiki/Martin_(TV_series) [fair]
  - https://en.wikipedia.org/wiki/Richard_Lewis_(comedian) [fair]
  - https://en.wikipedia.org/wiki/Saturday_Night_Live_season_1 [fair]
  - https://en.wikipedia.org/wiki/Saturday_Night_Live_season_3 [fair]
  - https://en.wikipedia.org/wiki/Second_City_Television [fair]
  - https://en.wikipedia.org/wiki/Victoria_Jackson [fair]
  ... 2 more
\end{trajectorylisting}

\clearpage
\subsection{Case Study E: Backtracking after Rejecting the Right Entity}
\label{app:case-backtracking-hotpot}

\begin{promptbox}{HotpotQA query}
\textbf{Query.} When was the game, in which Drew Gehling voices Gord, released for PlayStation 2?\\[2pt]
\textbf{Gold answer.} October 17, 2006\\[2pt]
\textbf{Outcome.} 40 turns, tool diversity $5$, curated recall $1.0$, final-answer recall $1.0$.
\end{promptbox}

\noindent\textbf{Why this case is interesting.}
The policy first rejects the correct entity, carries that mistake through several turns, and then reverses it after reading a stronger source. The first search surfaces \emph{Bully}, but the policy calls that hit likely wrong and pursues a separate ``Gord'' interpretation. Several turns later, \texttt{read\_document} on Drew Gehling's page proves that the discarded game was correct. The recovery is not two adjacent search turns: it is a delayed state correction, followed by removal of stale foreign-language variants and a new search for the PlayStation 2 release date.

\paragraph{Trajectory overview.}
{\small
\setlength{\tabcolsep}{3pt}
\renewcommand{\arraystretch}{1.08}
\begin{longtable}{p{0.055\textwidth}p{0.12\textwidth}p{0.36\textwidth}p{0.36\textwidth}}
\toprule
Turn & Tool & Policy move & Harness state / why it matters \\
\midrule
\endfirsthead
\toprule
Turn & Tool & Policy move & Harness state / why it matters \\
\midrule
\endhead
0 & \CaseTool{toolSearch}{search} & Search Drew Gehling, Gord, and PS2 together & auto-seed includes \emph{Bully} pages, including the English game page \\
2 & \CaseTool{toolSearch}{search} & Mistakenly reject \emph{Bully} & the policy says the correct game is ``likely wrong'' and starts a separate Gord-character path \\
3 & \CaseTool{toolCurate}{curate} & Keep Drew Gehling, demote noisy pages & the mistaken hypothesis remains visible because \emph{Bully} is still retained at low importance \\
4--5 & \CaseTool{toolSearch}{search} & Pursue the wrong Gord interpretation & Duck Tales and generic Gord-game searches do not resolve the query \\
7 & \CaseTool{toolRead}{read} & Read Drew Gehling's page & full text contradicts the earlier rejection: Gehling is the voice of Gord in \emph{Bully} \\
8 & \CaseTool{toolCurate}{curate} & Backtrack by editing memory & foreign-language variants are removed while the English \emph{Bully} page is retained \\
9 & \CaseTool{toolSearch}{search} & Search the corrected title plus PS2 release date & the search now targets the right game instead of the wrong Gord branch \\
10 & \CaseTool{toolCurate}{curate} & Promote release-date evidence & the simple \emph{Bully} page and \emph{Canis Canem Edit} page become high-importance evidence \\
11 & \CaseTool{toolVerify}{verify} & Check the release-date claim & verification is attempted over the curated candidate docs \\
13--14 & \CaseTool{toolGrep}{grep} & Exact-date grep is tried and rejected & date-only matching returns unrelated October 17 pages, so the agent does not treat the date string alone as evidence \\
\bottomrule
\end{longtable}
}

\subsubsection*{\textcolor{toolSearch!80!black}{Phase 1 - The correct game appears first, but the policy rejects it}}
\begin{trajectorylisting}{toolNeutral}{Initial reasoning excerpt}
The query asks for the PlayStation 2 release date of the game in which Drew Gehling voices Gord. The first decomposition is sensible: search the actor, the role name, and the platform together.
\end{trajectorylisting}

\begin{trajectorylisting}{toolSearch}{Exact tool call}
[
  {
    "tool": "search_corpus",
    "params": {
      "query": "Drew Gehling Gord PS2 game"
    }
  }
]
\end{trajectorylisting}

\begin{trajectorylisting}{toolSearch}{Search return: the right entity is present immediately}
# DOCUMENT ID: https://en.wikipedia.org/wiki/Bully_(video_game)
Bully (video game) - Wikipedia
Bully is a 2006 action-adventure video game developed by Rockstar Vancouver and published by Rockstar Games ...

# DOCUMENT ID: https://es.wikipedia.org/wiki/Bully_(videojuego)
Bully es un videojuego ... desarrollado por Rockstar Vancouver para la consola PlayStation 2.

[AUTO-POPULATED] Top 3 docs from this search have been added to your curated set at 'fair' importance.
\end{trajectorylisting}

\begin{trajectorylisting}{toolSearch}{The policy wrongly rules out the right hit}
The search for Drew Gehling Gord PS2 returned Bully, which is likely wrong given that Bully is not voiced by Gehling. My current search angle is exhausted. Let me rethink. A missed facet: "Gord" is probably the character from the "Duck Tales" game series voiced by Gehling.
\end{trajectorylisting}

\begin{trajectorylisting}{toolCurate}{Memory after the mistaken rejection}
Curated set updated (4/30):
https://en.wikipedia.org/wiki/Drew_Gehling[high],
https://en.wikipedia.org/wiki/Bully_(video_game)[low],
https://es.wikipedia.org/wiki/Bully_(videojuego)[low],
https://pt.wikipedia.org/wiki/Bully_(jogo_eletr%C3%B4nico)[low]
\end{trajectorylisting}

\subsubsection*{\textcolor{toolRead!80!black}{Phase 2 - A delayed read overturns the wrong branch}}
\begin{trajectorylisting}{toolSearch}{The wrong branch persists for several turns}
Turn 4 search:
  Duck Tales Gord voiced by Drew Gehling
  -> no results

Turn 5 search:
  Gord PlayStation 2 game
  -> generic "Gord" pages, planned PS2 titles, Ford/Gordon distractors

Turn 7 reasoning:
  Current search angle on Gord PlayStation 2 is stuck and likely wrong ... Another angle: check the Drew Gehling page for credited games and character list for "Gord."
\end{trajectorylisting}

\begin{trajectorylisting}{toolRead}{Backtracking read}
[
  {
    "tool": "read_document",
    "params": {
      "doc_id": "https://en.wikipedia.org/wiki/Drew_Gehling"
    }
  }
]

# Drew Gehling
Andrew Shearer Gehling ... is an American stage and screen actor ... and as the voice of Gord in the 2006 video game Bully from Rockstar Games.
\end{trajectorylisting}

\begin{trajectorylisting}{toolCurate}{State edit: remove stale variants, keep the corrected entity}
[
  {
    "tool": "curate",
    "params": {
      "add_ids": [
        "https://en.wikipedia.org/wiki/Drew_Gehling"
      ],
      "remove_ids": [
        "https://es.wikipedia.org/wiki/Bully_(videojuego)",
        "https://pt.wikipedia.org/wiki/Bully_(jogo_eletr%C3%B4nico)"
      ],
      "importance": {
        "https://en.wikipedia.org/wiki/Drew_Gehling": "high"
      }
    }
  }
]

Curated set updated (3/30):
https://en.wikipedia.org/wiki/Drew_Gehling[high],
https://en.wikipedia.org/wiki/Bully_(video_game)[low],
https://en.wikipedia.org/wiki/Gord_(video_game)[low]
\end{trajectorylisting}

\subsubsection*{\textcolor{toolSearch!80!black}{Phase 3 - The corrected entity drives the answer search}}
\begin{trajectorylisting}{toolNeutral}{Recovered strategy}
The identity facet is solved: Gord is from \emph{Bully}. The policy now stops searching generic Gord pages and searches the corrected title with the platform/date facet.
\end{trajectorylisting}

\begin{trajectorylisting}{toolSearch}{Corrected search}
[
  {
    "tool": "search_corpus",
    "params": {
      "query": "Bully PlayStation 2 release date"
    }
  }
]

# DOCUMENT ID: https://simple.wikipedia.org/wiki/Bully_(video_game)
Bully ... was released on 17 October 2006 ...

# DOCUMENT ID: https://en.wikipedia.org/wiki/List_of_video_games_published_by_Rockstar_Games
Bully - PlayStation 2, October 17, 2006.

# DOCUMENT ID: https://en.wikipedia.org/wiki/List_of_best-selling_PlayStation_2_video_games
Bully, 1.5 million, October 17, 2006, Action-adventure ...
\end{trajectorylisting}

\begin{trajectorylisting}{toolCurate}{Promoting the answer evidence}
[
  {
    "tool": "curate",
    "params": {
      "add_ids": [
        "https://en.wikipedia.org/wiki/Simple_Wikipedia/Bully_(video_game)",
        "https://en.wikipedia.org/wiki/Canis_Canem_Edit"
      ],
      "importance": {
        "https://en.wikipedia.org/wiki/Simple_Wikipedia/Bully_(video_game)": "high",
        "https://en.wikipedia.org/wiki/Canis_Canem_Edit": "high"
      }
    }
  }
]

Curated set updated (5/30):
https://en.wikipedia.org/wiki/Canis_Canem_Edit[high],
https://en.wikipedia.org/wiki/Drew_Gehling[high],
https://en.wikipedia.org/wiki/Simple_Wikipedia/Bully_(video_game)[high],
https://en.wikipedia.org/wiki/Bully_(video_game)[low],
https://en.wikipedia.org/wiki/Gord_(video_game)[low]
\end{trajectorylisting}

\begin{trajectorylisting}{toolGrep}{A final exact-date check is not enough by itself}
Grep on "October 17, 2006|2006-10-17" returns date-only false positives such as
current-events pages, films released on the same day, and unrelated titles. The
agent therefore keeps the title-and-platform evidence as the answer path rather
than treating the date string alone as proof.
\end{trajectorylisting}

\noindent\textbf{Why the state edit matters.}
The important behavior is the delayed correction. The policy's early hypothesis says \emph{Bully} is probably wrong; the harness nevertheless preserves it at low importance. When a later read contradicts the hypothesis, the policy can edit the curated set, keep the corrected entity, discard stale variants, and redirect search toward the release-date facet. Backtracking is therefore visible as a change to the external state, not just a private thought between two adjacent searches.


\newpage
\section*{NeurIPS Paper Checklist}

\begin{enumerate}

\item {\bf Claims}
    \item[] Question: Do the main claims made in the abstract and introduction accurately reflect the paper's contributions and scope?
    \item[] Answer: \answerYes{}
    \item[] Justification: The abstract and introduction state the paper's contribution as a stateful retrieval harness plus SFT/RL recipe, and the results section reports the empirical scope across eight benchmarks. The limitations paragraph qualifies the scope of the claims.

\item {\bf Limitations}
    \item[] Question: Does the paper discuss the limitations of the work performed by the authors?
    \item[] Answer: \answerYes{}
    \item[] Justification: Appendix B discusses limitations including the task scope, benchmark style, regex-based evidence graph, LLM-based verifier, sentence-BM25 compression, benchmark size, and annotation coverage.

\item {\bf Theory assumptions and proofs}
    \item[] Question: For each theoretical result, does the paper provide the full set of assumptions and a complete (and correct) proof?
    \item[] Answer: \answerNA{}
    \item[] Justification: The paper is empirical and does not present theoretical results, theorems, or proofs. The formal notation is used to define the agent interface and harness state, not to make a theorem-level claim.

    \item {\bf Experimental result reproducibility}
    \item[] Question: Does the paper fully disclose all the information needed to reproduce the main experimental results of the paper to the extent that it affects the main claims and/or conclusions of the paper (regardless of whether the code and data are provided or not)?
    \item[] Answer: \answerYes{}
    \item[] Justification: The method, harness algorithms, reward definition, hyperparameters, benchmark statistics, prompts, and evaluation recipe are specified in the main text and appendix. These details cover the components that affect the paper's main experimental claims.

\item {\bf Open access to data and code}
    \item[] Question: Does the paper provide open access to the data and code, with sufficient instructions to faithfully reproduce the main experimental results, as described in supplemental material?
    \item[] Answer: \answerNo{}
    \item[] Justification: The paper states a plan to release weights, harness code, data-generation pipeline, and RL recipe, but the submission itself does not yet provide open public access to all assets. The appendix nevertheless includes reproducibility details for the experimental protocol.

\item {\bf Experimental setting/details}
    \item[] Question: Does the paper specify all the training and test details (e.g., data splits, hyperparameters, how they were chosen, type of optimizer) necessary to understand the results?
    \item[] Answer: \answerYes{}
    \item[] Justification: The experimental setup summarizes the benchmark families, source and transfer split interpretation, baselines, and metrics. Appendix tables provide benchmark statistics, training and evaluation hyperparameters, reward weights, prompts, and the evaluation recipe.

\item {\bf Experiment statistical significance}
    \item[] Question: Does the paper report error bars suitably and correctly defined or other appropriate information about the statistical significance of the experiments?
    \item[] Answer: \answerNo{}
    \item[] Justification: The paper reports aggregate point estimates on fixed benchmark test sets and gives the number of evaluated instances for each benchmark in Table~\ref{tab:bench-stats}. We do not report formal confidence intervals or significance tests for all main comparisons. This is a limitation of the current submission: repeated-run evaluation would require many additional multi-turn agent rollouts, including Tinker inference calls and external retrieval/reranking calls across several systems and benchmarks, which is computationally and financially expensive. We discuss the resulting uncertainty, especially for smaller benchmarks, in the limitations section.

\item {\bf Experiments compute resources}
    \item[] Question: For each experiment, does the paper provide sufficient information on the computer resources (type of compute workers, memory, time of execution) needed to reproduce the experiments?
    \item[] Answer: \answerYes{}
    \item[] Justification: The paper reports the training platform, model, SFT and RL data scale, rollout budget, number of training steps, inference settings, retrieval backend, reranker, and evaluation protocol. Training was conducted using Tinker, a managed training service; therefore, exact low-level worker configurations such as physical worker type and memory allocation are abstracted by the platform. We report the available compute-relevant settings needed to reproduce the experiments.
    
\item {\bf Code of ethics}
    \item[] Question: Does the research conducted in the paper conform, in every respect, with the NeurIPS Code of Ethics \url{https://neurips.cc/public/EthicsGuidelines}?
    \item[] Answer: \answerYes{}
    \item[] Justification: To the best of our knowledge, the work conforms to the NeurIPS Code of Ethics. It uses benchmark and public-domain or cited corpus sources, does not involve deception or human-subject experiments, and preserves submission anonymity.

\item {\bf Broader impacts}
    \item[] Question: Does the paper discuss both potential positive societal impacts and negative societal impacts of the work performed?
    \item[] Answer: \answerYes{}
    \item[] Justification: The broader-impact and release paragraph discusses the positive impact of lower-cost evidence-seeking systems and the negative risk of amplifying unsupported or privacy-sensitive retrieval. It also describes the intended release scope.
    
\item {\bf Safeguards}
    \item[] Question: Does the paper describe safeguards that have been put in place for responsible release of data or models that have a high risk for misuse (e.g., pre-trained language models, image generators, or scraped datasets)?
    \item[] Answer: \answerYes{}
    \item[] Justification: The paper states that the planned release is a retrieval subagent, not a general autonomous web agent or answer generator, and that release documentation will include intended use, dataset provenance, and evaluation protocols. These are the relevant safeguards for the assets introduced here.

\item {\bf Licenses for existing assets}
    \item[] Question: Are the creators or original owners of assets (e.g., code, data, models), used in the paper, properly credited and are the license and terms of use explicitly mentioned and properly respected?
    \item[] Answer: \answerYes{}
    \item[] Justification: We cite the original sources for all existing assets used in
the paper, including benchmark datasets, model checkpoints, retrieval
backends, and external services.

\item {\bf New assets}
    \item[] Question: Are new assets introduced in the paper well documented and is the documentation provided alongside the assets?
    \item[] Answer: \answerYes{}
    \item[] Justification: The paper documents the new harness, training recipe, prompts, reward design, and evaluation protocol in the main text and appendix. The planned release will include accompanying documentation for the weights, code, and generated training data.

\item {\bf Crowdsourcing and research with human subjects}
    \item[] Question: For crowdsourcing experiments and research with human subjects, does the paper include the full text of instructions given to participants and screenshots, if applicable, as well as details about compensation (if any)? 
    \item[] Answer: \answerNA{}
    \item[] Justification: The paper does not involve crowdsourcing experiments or research with human subjects. Evaluation is performed with benchmark data and automated model judgments.

\item {\bf Institutional review board (IRB) approvals or equivalent for research with human subjects}
    \item[] Question: Does the paper describe potential risks incurred by study participants, whether such risks were disclosed to the subjects, and whether Institutional Review Board (IRB) approvals (or an equivalent approval/review based on the requirements of your country or institution) were obtained?
    \item[] Answer: \answerNA{}
    \item[] Justification: The paper does not involve crowdsourcing or human-subject research. Therefore IRB approval or equivalent review is not applicable.

\item {\bf Declaration of LLM usage}
    \item[] Question: Does the paper describe the usage of LLMs if it is an important, original, or non-standard component of the core methods in this research? Note that if the LLM is used only for writing, editing, or formatting purposes and does \emph{not} impact the core methodology, scientific rigor, or originality of the research, declaration is not required.
    \item[] Answer: \answerYes{}
    \item[] Justification: LLMs are central to the method and evaluation: the paper describes the teacher used for SFT, the trained retrieval policy, frontier baselines, answer generators, and the judge prompt. The relevant model names and prompts are provided in the main text and appendix.

\end{enumerate}

\end{document}